\documentclass[11pt,a4paper]{article}
\usepackage[dvipsnames]{xcolor}
\usepackage[hyperref]{emnlp2020}
\usepackage{times}
\usepackage{latexsym}
\usepackage{graphicx}
\usepackage{booktabs}
\usepackage{subcaption}
\usepackage{amsmath}
\usepackage{multirow}
\usepackage{comment}
\usepackage{algorithmic}
\usepackage{graphicx}
\usepackage{caption,subcaption}
\usepackage[ruled,vlined]{algorithm2e}
\usepackage{listings}
\usepackage{amssymb}
\usepackage{pifont}
\usepackage{float}
\usepackage[export]{adjustbox}
\usepackage{caption}
\definecolor{blue1}{rgb}{0.96, 0.74, 0.75}
\definecolor{blue2}{rgb}{0.94, 0.45, 0.43}
\definecolor{blue4}{rgb}{0.92, 0.3, 0.27}
\definecolor{blue3}{rgb}{0.86, 0.09, 0.07}
\definecolor{bostonuniversityred}{rgb}{0.8, 0.0, 0.0}
\definecolor{blue(pigment)}{rgb}{0.2, 0.2, 0.6}

\definecolor{blue1}{rgb}{0.99,0.9,0.9}
\definecolor{blue2}{rgb}{0.99,0.7,0.7}
\definecolor{blue3}{rgb}{0.99,0.5,0.5}

\definecolor{blue_graph}{rgb}{0.72, 0.80, 0.98}
\definecolor{green_graph}{rgb}{0.8, 0.95, 0.8}
\definecolor{red_graph}{rgb}{0.98, 0.8, 0.8}
\definecolor{green}{rgb}{0.8, 0.95, 0.8}
\usepackage{xcolor}
\usepackage{graphicx}
\newcommand{\hz}{\vphantom{\parbox[c]{0.1cm}{\rule{0.1cm}{0.4cm}}}}

\newcommand{\cmark}{\color{gray!90}\ding{51}}%
\newcommand{\xmark}{\ding{55}}%
\def\u{\underline }
\usepackage{microtype}
\usepackage{xspace}
\usepackage{paralist}
\usepackage{xcolor}
\newcommand{\ensuretext}[1]{#1}
\newcommand{\mcmarker}{\ensuretext{\textcolor{magenta}{\ensuremath{^{\textsc{M}}_{\textsc{C}}}}}}

\newcommand{\mycomment}[3]{}
\newcommand{\mc}[1]{\mycomment{\mcmarker}{#1}{magenta}}

\newcommand{\ignore}[1]{}
\newcommand{\modelname}{Divergent m\textsc{bert}\xspace}
\newcommand{\pos}{\textsc{eq}\xspace}
\newcommand{\negat}{\textsc{div}\xspace}
\newcommand{\dataset}{\textsc{refresd}\xspace}
\newcommand{\negatend}{\textsc{div}}
\aclfinalcopy 

\title{Ranking Semantic Divergence Intensity For Better Detection in Parallel Sentences with Minimum Supervision}
\title{Detecting Fine-Grained Cross-Lingual Semantic Divergences \\ without Supervision by Learning to Rank}

\author{Eleftheria Briakou \normalfont{and}  \textbf{Marine Carpuat} \\
  Department of Computer Science \\
  University of Maryland\\
  College Park, MD $20742$, USA\\
  \texttt{\href{mailto:ebriakou@cs.umd.edu}{ebriakou@cs.umd.edu}, \href{mailto:marine@cs.umd.edu}{marine@cs.umd.edu}}} 

\date{}

\begin{document}
\maketitle

\begin{abstract}

Detecting fine-grained differences in content conveyed in different languages matters for cross-lingual \textsc{nlp} and multilingual corpora analysis, but it is a challenging machine learning problem since annotation is expensive and hard to scale.~This work improves the prediction and annotation of fine-grained semantic divergences.~We introduce a training strategy for multilingual \textsc{bert} models by learning to rank synthetic divergent examples of varying granularity.~We evaluate our models on the~\textbf{R}ationalized~\textbf{E}nglish-\textbf{Fre}nch~\textbf{S}emantic~\textbf{D}ivergences, a new dataset released with this work, consisting of English-French sentence-pairs annotated with semantic divergence classes and token-level rationales.~Learning to rank helps detect fine-grained sentence-level divergences more accurately than a strong sentence-level similarity model, while token-level predictions have the potential of further distinguishing between coarse and fine-grained divergences.

\end{abstract}

\section{Introduction}

Comparing and contrasting the meaning of text conveyed in different languages is a fundamental \textsc{nlp} task. It can be used to curate clean parallel corpora for downstream tasks such as machine translation~\citep{koehn-etal-2018-findings}, cross-lingual transfer learning, or semantic modeling~\citep{ganitkevitch-etal-2013-ppdb,lample2019large}, and it is also useful to directly analyze multilingual corpora. For instance, detecting the commonalities and divergences between sentences drawn from English and French Wikipedia articles about the same topic would help analyze language bias~\citep{BaoHechtCartonQuaderiHornGergle2012,MassaScrinzi2012}, or mitigate differences in coverage and usage across languages~\citep{YeungDuhNagata2011,WulczynWestZiaLeskovec2016,LemmerichSaez-TrumperWestZia2019}. This requires not only detecting coarse content mismatches, but also fine-grained differences in sentences that overlap in content. 
Consider the following English and French sentences, sampled from the WikiMatrix parallel corpus. While they share important content, highlighted words convey meaning missing from the other language:

\begin{tabular}{p{0.9\columnwidth}}
\\
\textsc{en} \it Alexander Muir's ``The Maple Leaf Forever'' \textbf{served for many years} as an \textbf{unofficial} Canadian \textbf{national anthem}.\\
\textsc{fr} \it Alexander Muir compose The Maple Leaf Forever (en) qui est un \textbf{chant patriotique} pro canadien \textbf{anglais}.\\
\textcolor{darkgray}{\textsc{gloss} \it Alexander Muir composes The Maple Leaf Forever which is an English Canadian patriotic song.}\\
\\
\end{tabular}

We show that explicitly considering diverse types of semantic divergences in bilingual text benefits both the annotation and prediction of cross-lingual semantic divergences. We create and release the \textbf{R}ationalized \textbf{E}nglish-\textbf{Fre}nch \textbf{S}emantic \textbf{D}ivergences corpus (\dataset), based on a novel divergence annotation protocol that exploits rationales to improve annotator agreement. We introduce \modelname, a  \textsc{bert}-based model that detects fine-grained semantic divergences without supervision by learning to rank synthetic divergences of varying granularity. Experiments on \dataset show that our model distinguishes semantically equivalent from divergent examples much better than a strong sentence similarity baseline and that unsupervised token-level divergence tagging offers promise to refine distinctions among divergent instances. We make our code and data publicly available.\footnote{Implementations of \modelname can be found at: \url{https://github.com/Elbria/xling-SemDiv}; the \dataset dataset is hosted at: 
 \url{https://github.com/Elbria/xling-SemDiv/tree/master/REFreSD}.}

\section{Background}

Following \citet{Vyas2018IdentifyingSD}, we use the term \textbf{cross-lingual semantic divergences} to refer to differences in meaning between sentences written in two languages. Semantic divergences differ from typological divergences that reflect different ways of encoding the same information across languages \cite{dorr-1994-machine}. 
In sentence pairs drawn from comparable documents---written independently in each language but sharing a topic---sentences that contain translated fragments are rarely exactly equivalent \cite{fung-cheung-2004-multi,Munteanu:2005:IMT:1110825.1110828}, and sentence alignment errors yield coarse mismatches in meaning \cite{Goutte2012TheIO}.
In translated sentence pairs, differences in discourse structure across languages \citep{li-etal-2014-cross} can lead to sentence-level divergences or discrepancies in translation of pronouns
\cite{lapshinova-koltunski-hardmeier-2017-discovery,sostaric-etal-2018-discourse};  translation lexical choice requires selecting between near synonyms that introduce language-specific nuances \cite{Hirst1995NearsynonymyAT}; typological  divergences  lead  to  structural mismatches \cite{dorr-1994-machine}, and non-literal translation processes can lead to semantic drifts \cite{zhai-etal-2018-construction}. 

Despite this broad spectrum of phenomena, recent work has effectively focused on coarse-grained divergences: \citet{Vyas2018IdentifyingSD} work on subtitles and Common Crawl corpora where sentence alignment errors abound, and \citet{pham-etal-2018-fixing} focus on fixing divergences where content is appended to one side of a translation pair. 
By contrast, \citet{zhai-etal-2018-construction,Zhai2019TowardsRP} introduce token-level annotations that capture the meaning changes introduced by human translators during the translation process \citep{MolinaHurtadoAlbir2002}. However, this expensive annotation process does not scale easily.

When processing bilingual corpora, any meaning mismatches between the two languages are primarily viewed as noise for the downstream task. In shared tasks for filtering web-crawled parallel corpora \citep{koehn-etal-2018-findings,koehn-etal-2019-findings}, the best performing systems rely on 
translation models, or cross-lingual sentence embeddings to place bilingual sentences on a clean to noisy scale \cite{junczys-dowmunt-2018-dual, sanchez-cartagena-etal-2018-prompsits, lu-etal-2018-alibaba, DBLP:journals/corr/abs-1906-08885}. When mining parallel segments in Wikipedia for the WikiMatrix corpus \cite{DBLP:journals/corr/abs-1907-05791},  examples are ranked using the \textsc{laser} score \cite{artetxe2019tacl}, which computes cross-lingual similarity in a language-agnostic sentence embedding space. While this approach yields a very useful corpus of $135$M parallel sentences in $1{,}620$ language pairs, we show that \textsc{laser} fails to detect many semantic divergences in WikiMatrix.

\section{Unsupervised Divergence Detection}

We introduce a model based on multilingual \textsc{bert} (m\textsc{bert}) to distinguish divergent from equivalent sentence-pairs (Section~\ref{sec:model}). In the absence of annotated training data, we derive synthetic divergent samples from parallel corpora (Section~\ref{synth_div}) and train via learning to rank to exploit the diversity and varying granularity of the resulting samples (Section~\ref{sec:training_strategy}). We also show how our model can be extended to label tokens within sentences (Section~\ref{sec:token_tagging}). 

\subsection{\modelname Model}\label{sec:model}

Following prior work~\citep{Vyas2018IdentifyingSD}, we frame divergence detection as binary classification
(equivalence vs.\ divergence) 
given two inputs: an English sentence $\mathbf{x_e}$ and a French sentence $\mathbf{x_f}$. Given the success of multilingual masked language models like m\textsc{bert}~\cite{devlin2018bert}, \textsc{xlm} \cite{lample2019large}, and \textsc{xlm-r}~\cite{conneau-etal-2020-unsupervised} on cross-lingual understanding tasks, we build our classifier on top of multilingual \textsc{bert} in a standard fashion: we create a sequence
$\mathbf{x}$ by concatenating $\mathbf{x_e}$ and $\mathbf{x_f}$ with helper delimiter tokens: $\mathbf{x} = (\textsc{[cls]},\mathbf{x_e},\textsc{[sep]},\mathbf{x_f},\textsc{[sep]})$. The \textsc{[cls]} token encoding serves as the representation for the sentence-pair $\mathbf{x}$, passed through a
feed-forward layer network $F$ to get the score $F(\mathbf{x})$. Finally, we convert the score $F(\mathbf{x})$ 
into the probability of $\mathbf{x}$ belonging to the equivalent class.

\subsection{Generating Synthetic Divergences}\label{synth_div}

\begin{figure*}[!ht]
    \centering
    \includegraphics[scale=0.33]{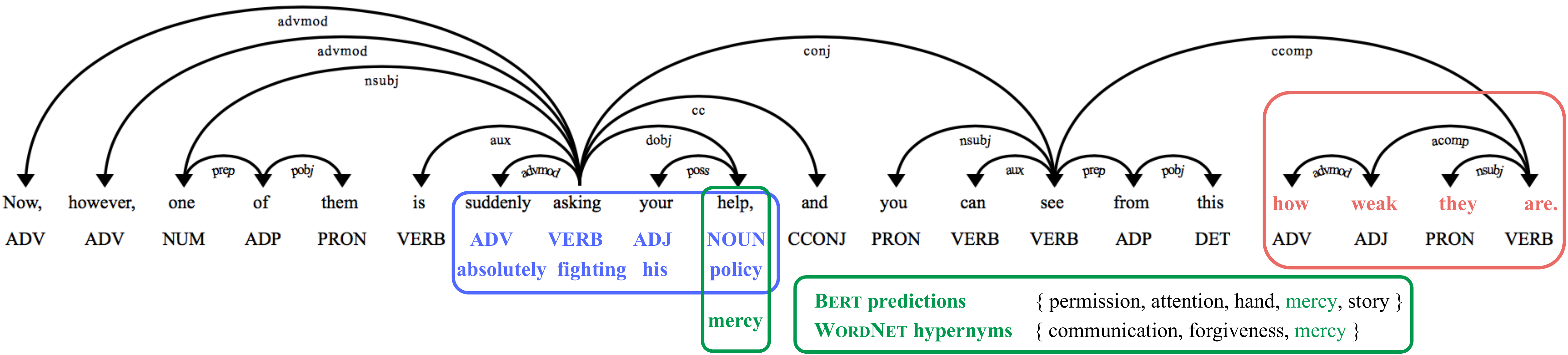}
    \label{fig:synthetic}
\end{figure*}
{
\renewcommand{\arraystretch}{1.}
\begin{table*}[!ht]
    \centering
    \scalebox{0.73}{
    \begin{tabular}{l}

    \textit{Seed Equivalent Sample} \\
    
    \hspace{2cm}  Now, however, one of them is suddenly asking your help, and you can see from this how weak they are. \\
    \hspace{2cm}  Maintenant, cependant, l'un d'eux vient soudainement demander votre aide et vous pouvez voir à quel point ils sont faibles.\\
    
   \addlinespace[0.02cm]\\
    \textit{Subtree Deletion} \\
    \hspace{2cm}  Now, however, one of them is suddenly asking your help, and you can see from this. \\
    \hspace{2cm}  Maintenant, cependant, l'un d'eux vient soudainement demander votre aide et vous pouvez voir à \colorbox{red_graph}{\hz quel} point \colorbox{red_graph}{\hz ils sont faibles}.\\

    \addlinespace[0.02cm]\\
    \textit{Phrase Replacement} \\
    \hspace{2cm}  Now, however, one of them is \colorbox{blue_graph}{\hz absolutely fighting his policy}, and you can see from this how weak they are. \\
    \hspace{2cm}  Maintenant, cependant, l'un d'eux vient \colorbox{blue_graph}{\hz soudainement demander votre aide} et vous pouvez voir à quel point ils sont faibles.\\
    
    \addlinespace[0.02cm]\\
     \textit{Lexical Substitution} \\
    \hspace{2cm}  Now, however, one of them is suddenly asking your \colorbox{green_graph}{\hz mercy}, and you can see from this. \\
    \hspace{2cm}  Maintenant, cependant, l'un d'eux vient soudainement demander votre \colorbox{green_graph}{\hz aide} et vous pouvez voir à quel point ils sont faibles.\\  

    \end{tabular}}
    \caption{Starting from a seed equivalent parallel sentence-pair, we create three types of divergent samples of varying granularity by introducing the highlighted edits.}
  \label{fig:synthetic_examples}
\end{table*}
}

We devise three ways of creating training instances that mimic divergences of varying granularity by perturbing seed equivalent samples from parallel corpora~(Table~\ref{fig:synthetic_examples}):

\paragraph{Subtree Deletion} We mimic semantic divergences due to content included only in one language by deleting a randomly selected subtree in the dependency parse of the English sentence, or French words aligned to English words in that subtree. We use subtrees that are not leaves, and that cover less than half of the sentence length. \newcite{duran, cardon-grabar-2020-reducing} successfully use this approach to compare sentences in the same language. 

\paragraph{Phrase Replacement} Following \citet{pham-etal-2018-fixing}, we introduce divergences that mimic phrasal edits or mistranslations by substituting random source or target sequences by another sequence of words with matching \textsc{pos} tags (to keep generated sentences as grammatical as possible).

\paragraph{Lexical Substitution} We mimic \textit{particularization} and \textit{generalization} translation operations~\citep{Zhai2019TowardsRP} by substituting English words with hypernyms or hyponyms from WordNet. The replacement word is the highest scoring WordNet candidate in context, according to a \textsc{bert} language model~\cite{zhou-etal-2019-bert, Qiang2019ASB}.

We call all these divergent examples \textbf{contrastive} because each divergent example contrasts with a specific equivalent sample from the seed set. The three sets of transformation rules above create divergences of varying granularity and create an \textbf{implicit ranking over divergent examples} based on the range of edit operations, starting from a single token with lexical substitution, to local short phrases for phrase replacement, and up to half the words in a sentence when deleting subtrees.

\subsection{Learning to Rank Contrastive Samples}\label{sec:training_strategy}

We train the \modelname model by learning to rank synthetic divergences. Instead of treating equivalent and divergent samples independently, we exploit their contrastive nature by explicitly pairing divergent samples with their seed equivalent sample when computing the loss. Intuitively, \textit{lexical substitution} samples should rank higher than \textit{phrase replacement} and \textit{subtree deletion} and lower than seed \textit{equivalents}:
we exploit this intuition by enforcing a margin between the scores of increasingly divergent samples.

Formally, let $\mathbf{x}$ denote an English-French sentence-pair and $\mathbf{y}$ a contrastive pair, with $\mathbf{x}>\mathbf{y}$ indicating that the divergence in $\mathbf{x}$ is finer-grained than in $\mathbf{y}$. For instance, we assume that  $\mathbf{x}>\mathbf{y}$ if $\mathbf{x}$ is generated by lexical substitution and $\mathbf{y}$ by subtree deletion.

At training time, given a set of contrastive pairs
$\mathcal{D} = \{(\mathbf{x},\mathbf{y})\}$, the model is trained to rank the score of the first instance higher than the latter by minimizing the following margin-based loss
\begin{equation}
    \small
    \begin{split} 
     \mathcal{L_{\mathrm{sent}}}  = \frac{1}{|\mathcal{D}|} \Big( \sum_{(\mathbf{x},\mathbf{y}) \in \mathcal{D}} \max \{0, \xi -F(\mathbf{x}) + F(\mathbf{y})\} \Big) 
    \end{split}
    \label{eq:margin}
\end{equation}
where $\xi$ is a hyperparameter margin that controls the score difference between the sentence-pairs $\mathbf{x}$ and $\mathbf{y}$.
This ranking loss has proved useful in supervised English semantic analysis tasks~\citep{li-etal-2019-learning}, and we show that it also helps with our cross-lingual synthetic data.

\subsection{\modelname for Token Tagging}\label{sec:token_tagging}

\begin{figure*}[!ht]
    \centering
    \includegraphics[scale=0.35]{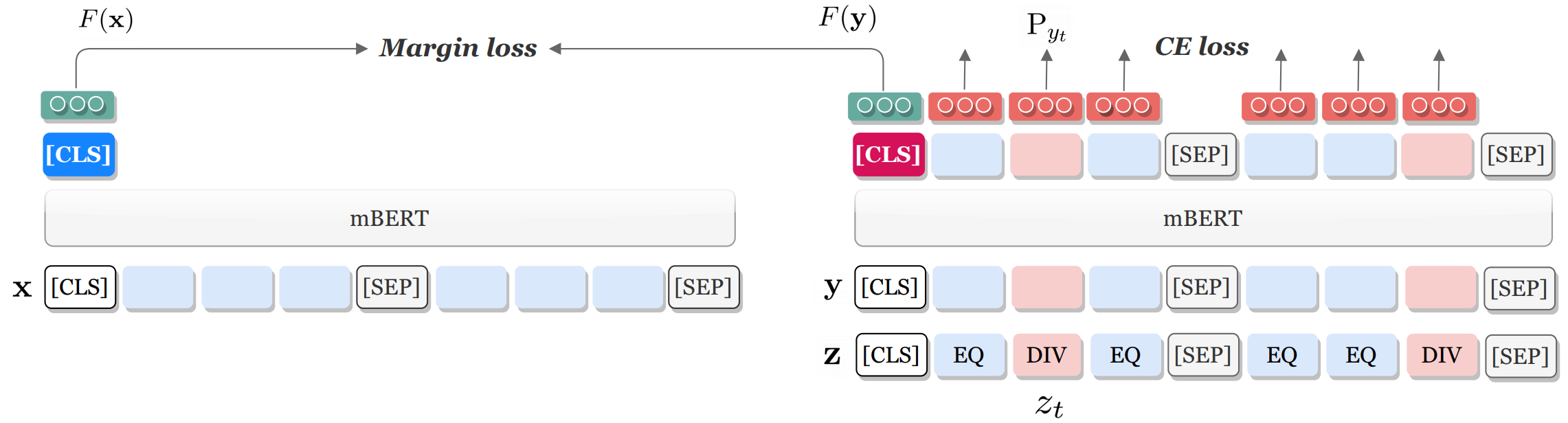}
    \caption{\modelname training strategy: given a triplet ($\mathbf{x}$,$\mathbf{y}$,$\mathbf{z}$), the model minimizes the sum of a margin-based loss via ranking a contrastive pair $\mathbf{x}>\mathbf{y}$ and a token-level cross-entropy loss on sequence labels $\mathbf{z}$.}
    \label{fig:model}
\end{figure*}

We introduce an extension of \modelname which, given a bilingual sentence pair, produces a) a sentence-level prediction of equivalence vs.\ divergence and b) a sequence of \pos/\negat labels for each input token. \pos and \negat refer to token-level tags of equivalence and divergence, respectively.

Motivated by annotation rationales, we adopt a multi-task framework
to train our model on a set of triplets $\mathcal{D'} = \{( \mathbf{x},\mathbf{y},\mathbf{z}) \}$,
still using only synthetic supervision (Figure~\ref{fig:model}). As in Section~\ref{sec:training_strategy}, we assume
 $\mathbf{x}>\mathbf{y}$, while $\mathbf{z}$ is the sequence of labels for the second encoded sentence pair $\mathbf{y}$, such that, at time $t$, $z_t\in$\{\pos,\negatend\} is the label of $y_t$. Since \modelname operates on sequences of subwords, we assign an \pos or \negat label to a word token if at leat one of its subword units is assigned that label.
 
 For the token prediction task, the final hidden state $h_t$ of each $y_t$ token is passed through a feed-forward layer and a softmax layer to produce the probability $\mathrm{P}_{y_t}$ of the $y_t$ token belonging to the \pos class. For the sentence task, the model learns to rank $\mathbf{x}>\mathbf{y}$, as in Section~\ref{sec:training_strategy}. We then minimize the sum of the sentence-level margin-loss and the average token-level cross-entropy loss ($\mathcal{L}_{\mathrm{\textsc{ce}}}$) across all tokens of $\mathbf{y}$, as defined in Equation \ref{loss}.
\begin{equation}
    \small
    \begin{split} 
     \mathcal{L}  = \frac{1}{|\mathcal{D'}|} \Big(  \sum_{(\mathbf{x},\mathbf{y}, \mathbf{z}) \in \mathcal{D'}} \big( \max \{0, \xi -F(\mathbf{x}) + F(\mathbf{y})\} \\
     + \frac{1}{|\mathbf{y}|} \sum_{t=1}^{|\mathbf{y}|} \mathcal{L}_{\mathrm{\textsc{ce}}} (\mathrm{P}_{y_t},  z_t)  \big) \Big) 
    \end{split}
    \label{loss}
\end{equation}
Similar multi-task models have been used for Machine Translation Quality Estimation~\citep{10.1145/3321127,kim-etal-2019-qe}, albeit with human-annotated training samples and a standard cross-entropy loss for both word-level and sentence-level sub-tasks.
\section{Rationalized English-French Semantic Divergences}\label{dataset_description}

\begin{figure}[H]
    \centering
\includegraphics[scale=0.27]{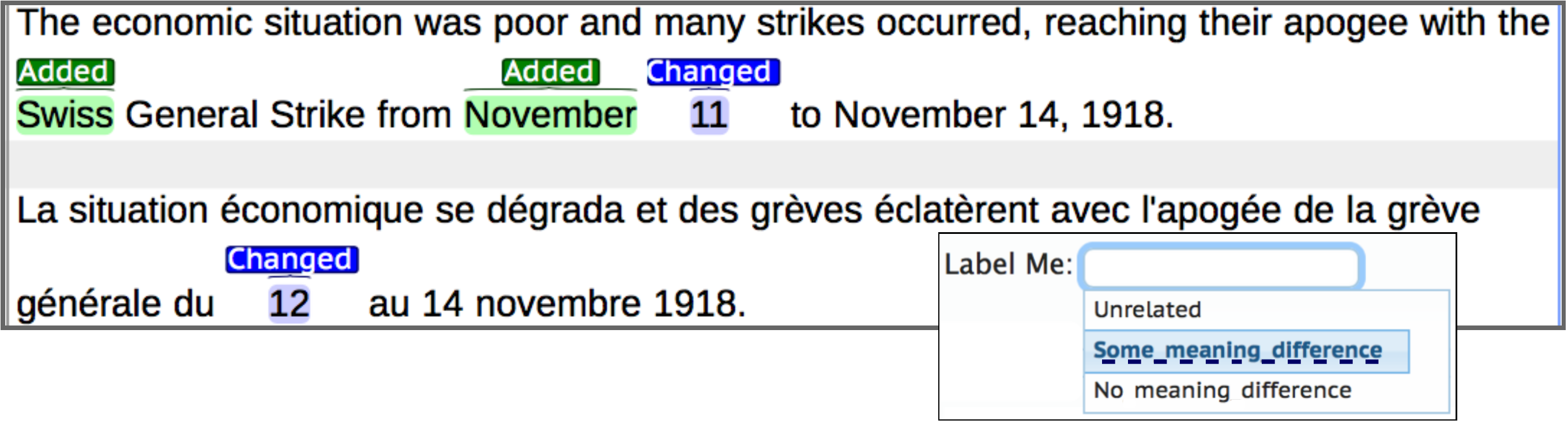}
    \caption{Screenshot of an example annotated instance.}
    \label{fig:interface}
\end{figure}

We introduce the \textbf{R}ationalized \textbf{E}nglish-\textbf{Fre}nch \textbf{S}emantic \textbf{D}ivergences (\dataset) dataset, which consists of $1{,}039$ English-French sentence-pairs annotated with sentence-level divergence judgments and token-level rationales. Figure~\ref{fig:interface} shows an example drawn from our corpus.

Our annotation protocol is designed to encourage annotators' sensitivity to semantic divergences other than misalignments, without requiring expert knowledge beyond competence in the languages of interest. We use two strategies for this purpose: (1)~we explicitly introduce distinct divergence categories for unrelated sentences and sentences that overlap in meaning; and (2)~we ask for annotation rationales~\cite{zaidan-eisner-piatko:2007:disc} by requiring annotators to highlight tokens indicative of meaning differences in each sentence-pair.  Thus, our approach strikes a balance between coarsely annotating sentences with binary distinctions that are fully based on annotators' intuitions~\cite{Vyas2018IdentifyingSD}, and exhaustively annotating all spans of a sentence-pair with fine-grained labels of translation processes~\cite{zhai-etal-2018-construction}.
We describe the annotation process 
and analysis of the collected instances based on data statements protocols described in \citet{bender-friedman-2018-data, datasheets}. We include more information in~\ref{sec:annotation_procedures}.

\paragraph{Task Description} An annotation instance consists of an English-French sentence-pair. Bilingual participants are asked to read them both and highlight tokens in each sentence that convey meaning not found in the other language. For each highlighted span, they pick whether this span conveys added information (``Added''),  information that is present in the other language but not an exact match (``Changed''), or some other type (``Other''). Those fine-grained classes are added to improve consistency across annotators and encourage them to read and compare the text closely. Finally, participants are asked to make a sentence-level judgment by selecting one of the following classes: ``No meaning difference'', ``Some meaning difference'', ``Unrelated''. Participants are not given specific instructions on how to use span annotations to make sentence-level decisions. Furthermore, participants have the option of using a text box to provide any comments or feedback on the example and their decisions. A summary of the different span and sentence labels along with the instructions given to participants can be found in~\ref{sec:annotation_guidelines}.

\paragraph{Curation rationale}\label{curation} Examples are drawn from the English-French section of the WikiMatrix corpus~\cite{DBLP:journals/corr/abs-1907-05791}. We choose this resource because (1) it is likely to contain diverse, interesting divergence types, since it consists of mined parallel sentences of diverse topics which are not necessarily generated by (human) translations, and (2) Wikipedia and WikiMatrix are widely used resources to train semantic representations and perform cross-lingual transfer in \textsc{nlp}. 
We exclude obviously noisy samples by filtering out sentence-pairs that a) are too short or too long, b) consist mostly of numbers, c) have a small token-level edit difference. The filtered version of the corpus consists of $2{,}437{,}108$ sentence-pairs.

\paragraph{Quality Control} We implement quality control strategies at every step. We build a dedicated task interface using the \textsc{brat} annotation toolkit~\cite{stenetorp-etal-2012-brat} (Figure~\ref{fig:interface}). We recruit participants from an educational institution and ensure they are proficient in both languages of interest. Specifically, participants are either bilingual speakers or graduate students pursuing a Translation Studies degree. We run a pilot study were participants annotate a sample containing both duplicated and reference sentence-pairs previously annotated by one of the authors. All annotators are found to be internally consistent on duplicated instances and agree with the reference annotations more than $60\%$ of the time. We solicit feedback from participants to finalize the instructions. 

\paragraph{Inter-annotator Agreement (\textsc{iaa})} We compute \textsc{iaa} for sentence-level annotations, as well as for the token and span-level rationales~(Table \ref{tab:iaaa}). 
We report $0.60$ Krippendorf's~$\alpha$ coefficient for sentence classes, which indicates a ``moderate'' agreement between annotators~\cite{Landis1977TheMO}. This constitutes a significant improvement over the $0.41$ and $0.49$ reported agreement coefficients on crowdsourced annotations of equivalence vs.\ divergence English-French parallel sentences drawn from OpenSubtitles and CommonCrawl corpora by prior work~\cite{Vyas2018IdentifyingSD}. 

Disagreements mainly occur between the ``No meaning difference'' and ``Some meaning difference'' classes, which we expect as different annotators might draw the line between which differences matter differently. 
We only observed $3$ examples where all $3$ annotators disagreed (tridisagreements), which indicates that the ``Unrelated'' and ``No meaning difference'' categories are more clear-cut. The rare instances with tridisagreements and bidisagreements---where the disagreement spans the two extreme classes---were excluded from the final dataset. Examples of \dataset corresponding to different levels of \textsc{iaa} are included in~\ref{sec:exampled_of_refresd}.

{
\renewcommand{\arraystretch}{1}
\begin{table}[H]
    \scalebox{0.95}{
    \centering
    \begin{tabular}{llc}
    \toprule[1.5pt]
    \textbf{Granularity} & \textbf{Method}     & \textbf{\textsc{iaa}}  \\
    \toprule[1pt]
    Sentence & Krippendorf's $\alpha$   &  $0.60$\\
    Span & macro F1 & $45.56 \pm 7.60$  \\
    Token & macro F1 & $33.94 \pm 8.24$\\
    \toprule[1.5pt]
    \end{tabular}}
    \caption{Inter-annotator agreement measured at different levels of granularities for the \dataset dataset.}
    \label{tab:iaaa}
\end{table}}

Quantifying agreement between rationales requires different metrics.~At the span-level, we compute macro F1 score for each sentence-pair following \citet{2019eraser}, where we treat one set of annotations as the reference standard and the other set as predictions. We count a prediction as a match if its token-level Intersection Over Union (\textsc{iou}) with any of the reference spans overlaps by more than some threshold (here, $0.5$). We report average span-level and token-level macro F1 scores, computed across all different pairs of annotators. Average statistics indicate that our annotation protocol enabled the collection of a high-quality dataset.

\paragraph{Dataset Statistics} Sentence-level annotations were aggregated by majority vote, yielding $252$, $418$, and $369$ instances for the ``Unrelated'', ``Some meaning difference'', and ``No meaning difference'' classes, respectively. In other words, 64\% of samples are divergent and 40\% of samples contain fine-grained meaning divergences, confirming that divergences vary in granularity and are too frequent to be ignored even in a corpus viewed as parallel.

\section{Experimental Setup}\label{sec:experimental_setup}

\paragraph{Data} We normalize English and French text in WikiMatrix consistently using the Moses toolkit~\cite{koehn-etal-2007-moses}, and tokenize into subword units using the ``BertTokenizer''. Specifically, our pre-processing pipeline consists of a) replacement of Unicode punctuation, b) normalization of punctuation, c) removing of non-printing characters, and d) tokenization.\footnote{ \url{https://github.com/facebookresearch/XLM/blob/master/tools/tokenize.sh}}
We align English to French bitext using the Berkeley word aligner.\footnote{\url{https://code.google.com/archive/p/berkeleyaligner}}
We filter out obviously noisy parallel sentences, as described in Section~\ref{dataset_description}, Curation Rationale. The top $5{,}500$ samples ranked by \textsc{laser} similarity score are treated as (noisy) equivalent samples and seed the generation of synthetic divergent examples.\footnote{ \url{https://github.com/facebookresearch/LASER/tree/master/tasks/WikiMatrix}} We split the seed set into $5{,}000$ training instances and $500$ development instances
consistently across experiments. Results on development sets for each experiment are included in~\ref{sec:dev_suppl}.

\paragraph{Models} Our models are based on the HuggingFace transformer library~\cite{Wolf2019HuggingFacesTS}.\footnote{\url{https://github.com/huggingface/transformers}} We fine-tune the ``\textsc{bert}-Base Multilingual Cased'' model~\cite{devlin2018bert},\footnote{\url{https://github.com/google-research/bert}} and perform a grid search on the margin hyperparameter, using the synthetic development set.  Further details on model and training settings can be found in \ref{app:implemantation_details}.

\paragraph{Evaluation} We evaluate all models on our new \dataset dataset using Precision, Recall, F1 for each class, and Weighted overall F1 score as computed by \textit{scikit-learn}~\cite{scikit-learn}.\footnote{\url{https://scikit-learn.org}} 

\section{Binary Divergence Detection}
\label{sec:binary}

We evaluate \modelname's ability to detect divergent sentence pairs in \dataset.

\subsection{Experimental Conditions}

\paragraph{\textsc{laser} baseline} This baseline distinguishes equivalent from divergent samples via a threshold on the \textsc{laser} score. We use the same threshold as \citet{DBLP:journals/corr/abs-1907-05791}, who show that training Neural Machine Translation systems on WikiMatrix samples with \textsc{laser} scores higher than $1.04$ improves \textsc{bleu}. Preliminary experiments suggest that tuning the \textsc{laser} threshold does not improve classification and that more complex models such as the \textsc{vdpwi} model used by \citet{Vyas2018IdentifyingSD} underperform \modelname, as discussed in~\ref{app:VDPWI}.

\paragraph{\modelname} We compare \modelname trained by learning to rank contrastive samples (Section~\ref{sec:training_strategy}) with ablation variants.

To test the impact of contrastive training samples, we fine-tune \modelname using 
\begin{inparaenum}
\item the Cross-Entropy (\textbf{\textsc{ce}}) loss on randomly selected synthetic divergences;
\item the \textbf{\textsc{ce}} loss on paired equivalent and divergent samples, treated as independent;
\item the proposed training strategy with a \textbf{Margin} loss to explicitly compare contrastive pairs. 
\end{inparaenum}

Given the fixed set of seed equivalent samples (Section~\ref{sec:experimental_setup}, Data), we vary the combinations of divergent samples: \begin{inparaenum}
\item \textbf{Single divergence type} we pair each seed equivalent with its corresponding divergent of that type, yielding a single contrastive pair;
\item \textbf{Balanced sampling} we randomly pair each seed equivalent with one of its corresponding divergent types, yielding a single contrastive pair;
\item \textbf{Concatenation} we pair each seed equivalent with one of each synthetic divergence type, yielding four contrastive pairs;
\item \textbf{Divergence ranking} we learn to rank pairs of close divergence types: equivalent vs.\ lexical substitution, lexical substitution vs.\ phrase replacement, or subtree deletion yielding four contrastive pairs.\footnote{We mimic both generalization \textbf{and} particularization.}
\end{inparaenum}

\subsection{Results}

All \modelname models outperform the \textsc{laser} baseline by a large margin (Table~\ref{tab:sentence_divergences_single}). The proposed training strategy performs best, improving over \textsc{laser}  by $31$ F1 points. Ablation experiments and analysis further show the benefits of diverse contrastive samples and learning to rank.

{
\setlength{\tabcolsep}{10pt}{
\renewcommand{\arraystretch}{1}
\begin{table*}[!ht]
    \centering
    \scalebox{0.85}{
    \begin{tabular}{llcrrrrrrrrr}
    \toprule[1.4pt]
     & & & \multicolumn{3}{c}{\textit{Equivalents}}& \multicolumn{3}{c}{\textit{Divergents}} & \multicolumn{3}{c}{\textit{All}} \\
    \cmidrule(lr){4-6}\cmidrule(lr){7-9}\cmidrule(lr){10-12}
    \textbf{Synthetic} & \textbf{Loss} & \textbf{Contrastive} & \textbf{P+} & \textbf{R+} & \textbf{F1+} & \textbf{P-} & \textbf{R-} & \textbf{F1-} &  \textbf{P} &  \textbf{R} & \textbf{F1}\\
    \addlinespace[0.20cm]
    \toprule[1pt]
    
    \addlinespace[0.20cm]
    \multirow{3}{*}{\textit{Phrase Replacement}} & \multirow{2}{*}{\textsc{ce}} & \xmark & $70$ & $56$ & $62$ & $78$ & $87$ & $82$ & $75$ & $76$ & $75$   \\
     & & \cmark &  $61$ & $81$ & $69$ & $87$ & $71$ & $78$ & $78$ & $75$ & $75$ \\
     & Margin & \cmark & $70$ & $76$ & $\u{73}$ & $86$ & $82$ & $\u{84}$ & $80$ & $80$ & $\u{80}$  \\ 

    \addlinespace[0.20cm]
    \multirow{3}{*}{\textit{Subtree Deletion}} & \multirow{2}{*}{\textsc{ce}} & \xmark & $81$ & $50$ & $62$ & $77$ & $93$ & $\u{85}$ & $78$ & $78$ & $77$       \\
    & & \cmark & $64$ & $84$ & $72$ & $89$ & $74$ & $81$ & $80$ & $77$ & $78$  \\
    & Margin  & \cmark & $70$ & $83$ & $\u{76}$ & $90$ & $81$ & $\u{85}$ & $83$ & $82$ & $\u{82}$ \\

    \addlinespace[0.20cm]
    \multirow{3}{*}{\textit{Lexical Substitution}} & \multirow{2}{*}{\textsc{ce}} & \xmark &  $65$ & $53$ & $57$ & $76$ & $84$ & $\u{80}$ & $72$ & $73$ & $\u{72}$  \\
    & & \cmark & $55$ & $81$ & $\u{66}$ & $86$ & $64$ & $73$ & $75$ & $70$ & $71$ \\
    & Margin & \cmark & $57$ & $75$ & $65$ & $83$ & $70$ & $76$ & $74$ & $72$ & $\u{72}$  \\
    
    \addlinespace[0.25cm]
    \hline
    
    \addlinespace[0.25cm]
    \multirow{3}{*}{\textit{Balanced}} & \multirow{2}{*}{\textsc{ce}} & \xmark & $76$ & $42$ & $54$ & $74$ & $93$ & $83$ & $75$ & $75$ & $73$ \\
    & & \cmark & $73$ & $73$ & $73$ & $85$ & $85$ & $85$ & $81$ & $81$ & $81$\\
    & Margin  & \cmark & $76$ & $73$ & $\u{75}$ & $85$ & $87$ & $\u{86}$ & $82$ & $82$ & $\u{82}$  \\
    
    \addlinespace[0.20cm]
    \multirow{3}{*}{\textit{Concatenation}} & \multirow{2}{*}{\textsc{ce}} & \xmark & $62$ & $32$ & $42$ & $70$ & $89$ & $79$ & $67$ & $69$ & $66$ \\
    & & \cmark & $73$ & $55$ & $63$ & $78$ & $89$ & $83$ & $76$ & $77$ & $76$ \\
    & Margin & \cmark & $84$ & $59$ & $\u{70}$ & $81$ & $94$ & $\u{87}$ & $82$ & $82$ & $\u{81}$ \\
  
    \addlinespace[0.20cm]
    \textit{Divergence Ranking} & Margin & \cmark & $82$ & $72$ & $\mathbf{77}$ & $86$ & $91$ & $\mathbf{88}$ & $84$ & $85$ & $\mathbf{84}$\\
    \addlinespace[0.25cm]
    \hline
    
    \addlinespace[0.25cm]
     \multicolumn{3}{l}{\textsc{laser} baseline} & $38$ & $58$ & $46$ & $68$ & $48$ & $57$ & $57$ & $52$ & $53$ \\
    \addlinespace[0.20cm]
    \toprule[1.4pt]
     \end{tabular}}
    \caption{Intrinsic evaluation of \modelname and its ablation variants on the \dataset dataset. We report Precision (P), Recall (R), and F1 for the equivalent (+) and divergent (-) classes separately, as well as for both classes (\textit{All}). \textit{Divergence Ranking} yields the best F1 scores across the board.}
    \label{tab:sentence_divergences_single}
\end{table*}
}}

\begin{figure*}[ht!]

    \begin{subfigure}{0.31\textwidth}
      \includegraphics[width=\linewidth]{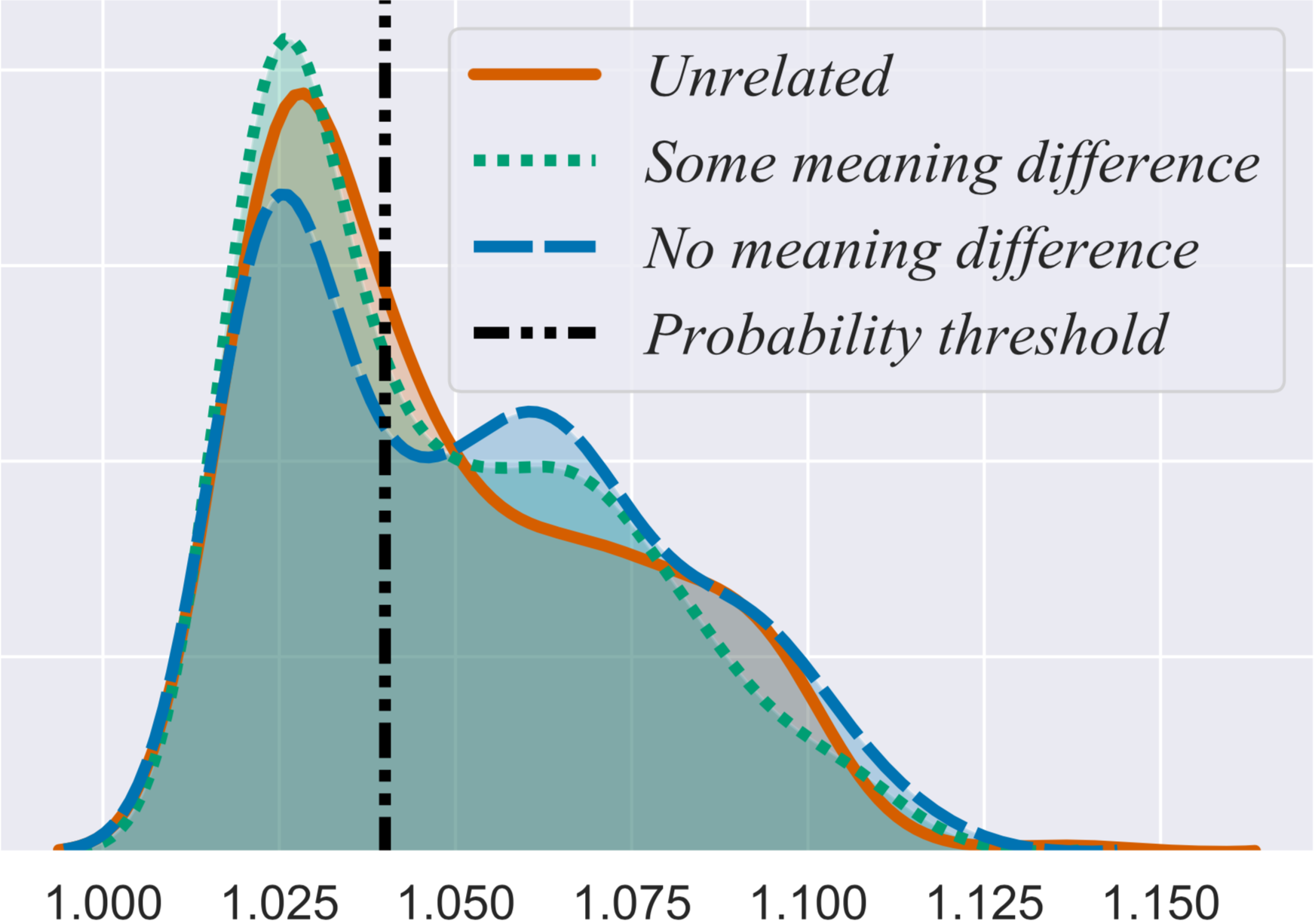}
      \caption{\textit{\textsc{laser}}} \label{fig:laser}
    \end{subfigure}\hfill
    \begin{subfigure}{0.31\textwidth}
      \includegraphics[width=\linewidth]{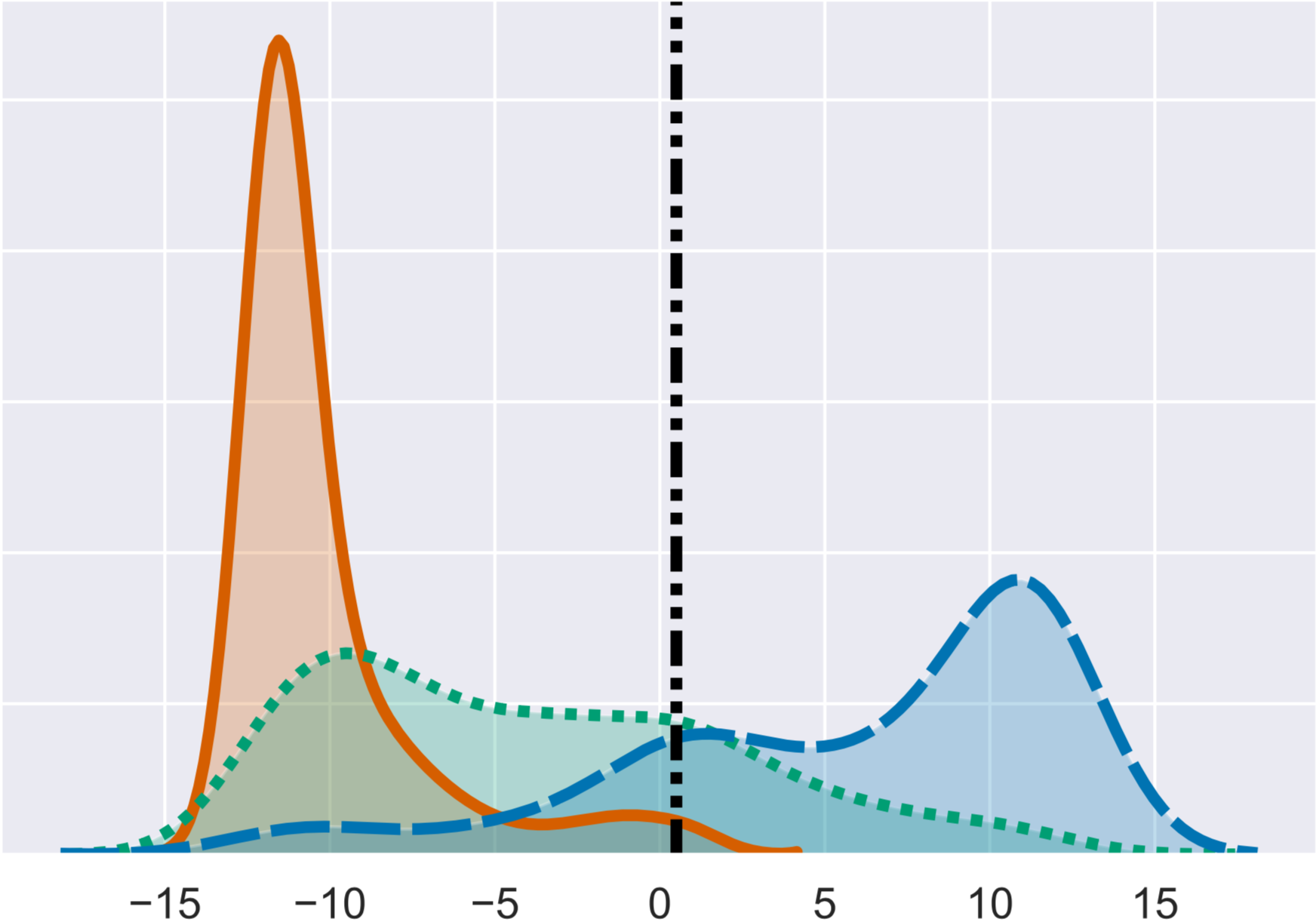}
         \caption{\textit{Subtree Deletion}}
    \end{subfigure}\hfill
    \begin{subfigure}{0.31\textwidth}
        \includegraphics[width=\linewidth]{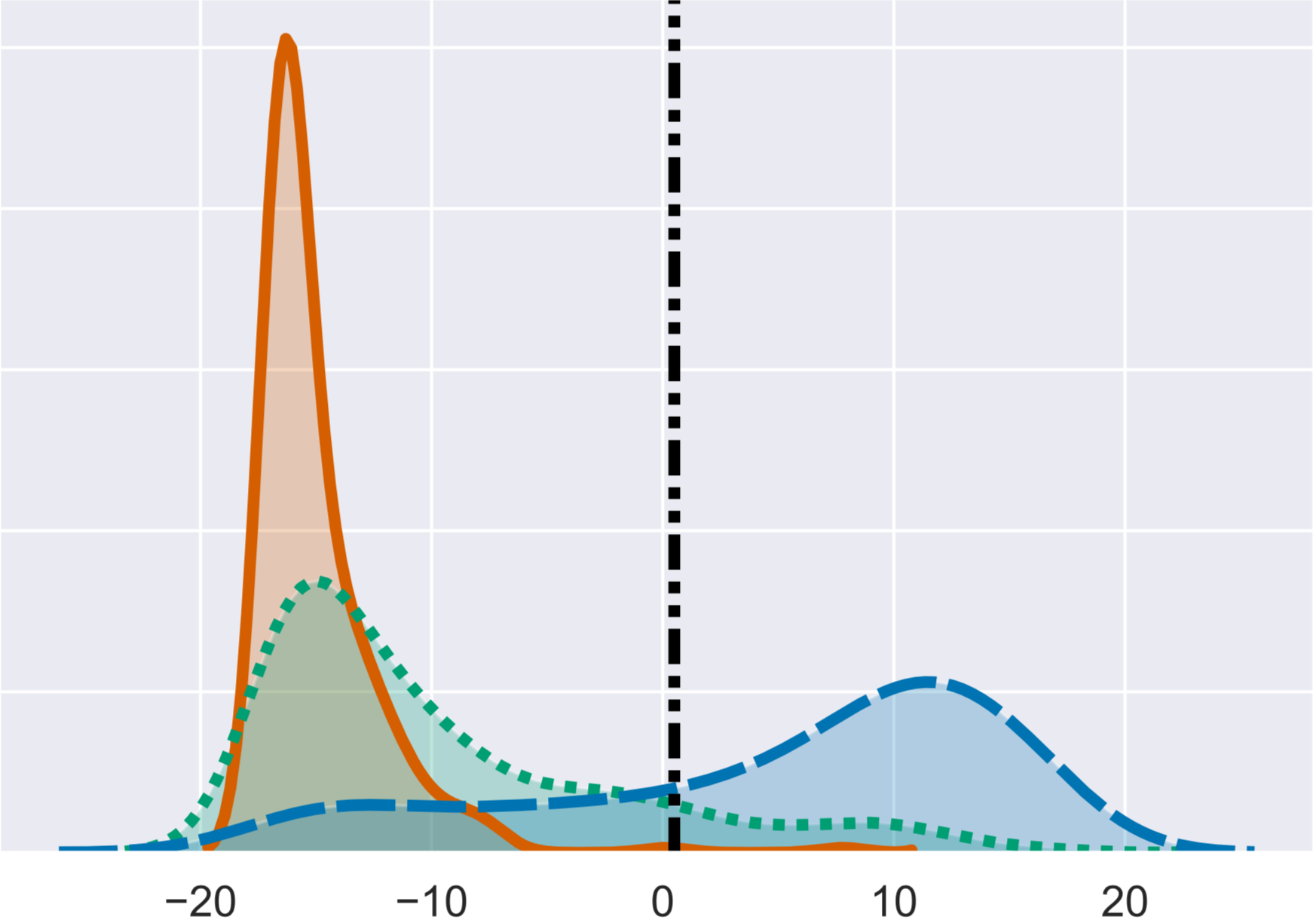}
       \caption{\textit{Divergence Ranking}}
    \end{subfigure}%

\caption{Score distributions assigned by different models to sentence-pairs of \dataset. \textit{Divergence Ranking} scores for the ``Some meaning difference'' class are correctly skewed more toward negative values.}\label{fig:distributions}
\end{figure*}

\paragraph{Contrastive Samples} With the \textsc{ce} loss, independent contrastive samples improve over randomly sampled synthetic instances overall ($+8.7$ F1+ points on average), at the cost of a smaller drop for the divergent class ($-5.3$ F1- points) for models trained on a single type of divergence. Using the margin loss helps models recover from this drop.

\paragraph{Divergence Types} All types improve over the \textsc{laser} baseline. When using a single divergence type, \textit{Subtree Deletion} performs best, even matching the overall F1 score of a system trained on all types of divergences (\textit{Balanced Sampling}). Training on the \textit{Concatenation} of all divergence types yields poor performance. We suspect that the model is overwhelmed by negative instances at training time, which biases it toward predicting the divergent class too often and hurting F1+ score for the equivalent class.

\paragraph{Divergence Ranking} How does divergence ranking improve predictions? Figure~\ref{fig:distributions} shows model score distributions for the $3$ classes annotated in \dataset. \textit{Divergence Ranking} particularly improves divergence predictions for the ``Some meaning difference'' class: the score distribution for this class is more skewed toward negative values than when training on contrastive \textit{Subtree Deletion} samples.

{ 
\setlength{\tabcolsep}{10pt}{
\renewcommand{\arraystretch}{1.1}
\begin{table*}[!ht]
    \centering
    \scalebox{0.69}{
    \begin{tabular}{lcccc@{0.1cm}ccc@{0.1cm}ccc}
    \toprule[1.4pt]
    & & \multicolumn{3}{c}{Union} & \multicolumn{3}{c}{Pair-wise Union}& \multicolumn{3}{c}{Intersection}\\
    \toprule[1pt]
    Model & Multi-task  & F1-\negat & F1-\pos & F1-Mul & F1-\negat & F1-\pos & F1-Mul & F1-\negat & F1-\pos & F1-Mul \\
    \toprule[0.5pt]

    \textit{Random Baseline} & & $0.21$ & $0.62$ & $0.13$ & $0.33$ & $0.59$ & $0.20$ & $0.21$ & $0.62$ & $0.13$ \\
    \textit{Token-only} &  & $0.39$ & $0.77$ & $0.30$ & $0.46$ & $\mathbf{0.88}$ & $0.41$ & $0.46$ & $\mathbf{0.92}$ & $0.42$\\

    \addlinespace[0.1cm]

    \textit{Balanced} & \cmark & $0.41$ & $0.77$ & $0.32$ & $0.46$ & $0.87$ & $0.40$ & $0.43$ & $0.91$ & $0.40$ \\
    \textit{Concatenation} & \cmark & $0.41$ & $\mathbf{0.78}$ & $0.32$ & $0.48$ & $\mathbf{0.88}$ & $0.42$ & $0.46$ & $\mathbf{0.92}$ & $0.42$  \\
    \textit{Divergence Ranking} & \cmark & $\mathbf{0.45}$ & $\mathbf{0.78}$ & $\mathbf{0.35}$ & $\mathbf{0.51}$ & $\mathbf{0.88}$ & $\mathbf{0.45}$ & $\mathbf{0.49}$ & $\mathbf{0.92}$ & $\mathbf{0.45}$ \\

    \addlinespace[0.2cm]

    \toprule[1.4pt]

    \end{tabular}}
    \caption{Evaluation of different models on the token-level prediction task for the \textbf{``Some meaning difference''} class of \dataset. \textit{Divergence Ranking} yields the best results across the board.
    }
    \label{tab:token-level-some_meanining-only}
\end{table*}
}}

\section{Finer-Grained Divergence Detection}\label{sec:Toward Finer-Grained Divergence Detection}

While we cast divergence detection as binary classification in Section~\ref{sec:binary}, human judges separated divergent samples into ``Unrelated'' and ``Some meaning difference'' classes in the \dataset dataset. Can we predict this distinction automatically? In the absence of annotated training data, we cannot cast this problem as a $3$-way classification, since it is not clear how the synthetic divergence types map to the $3$ classes of interest. 
Instead, we test the hypothesis that token-level divergence predictions can help discriminate between divergence granularities at the sentence-level, inspired by humans' use of rationales to ground sentence-level judgments.

\subsection{Experimental Conditions} 

\paragraph{Models} We fine-tune the \textbf{multi-task} m\textsc{bert} model that makes token and sentence predictions jointly, as described in Section~\ref{sec:token_tagging}. We contrast against a sequence labeling m\textsc{bert} model trained independently with the \textsc{ce} loss (\textbf{Token-only}). Finally, we run a \textbf{random baseline} where each token is labeled \pos or \negat uniformly at random.

\paragraph{Training Data} We tag tokens edited when generating synthetic divergences as \negat (e.g., highlighted tokens in Table~\ref{fig:synthetic_examples}), and others as \pos. Since edit operations are made on the English side, we tag aligned French tokens using the Berkeley aligner.

\paragraph{Evaluation} We expect token-level annotations in \dataset to be noisy since they are produced as rationales for sentence-level rather than token-level tags. We, therefore, consider three methods to aggregate rationales into token labels: a token is labeled as \negat if it is highlighted by at least one (\textbf{Union}), two (\textbf{Pair-wise Union}), or all three annotators (\textbf{Intersection}). We report F1 on the \negat and \pos class, and F1-Mul as their product for each of the three label aggregation methods.

\subsection{Results}

\paragraph{Token Labeling} We evaluate token labeling on \dataset samples from the ``Some meaning difference'' class, where we expect the more subtle differences in meaning to be found, and the token-level annotation to be most challenging (Table~\ref{tab:token-level-some_meanining-only}).
Examples of \modelname's token-level predictions are given in~\ref{app:token_predictions_example}. The \textit{Token-only} model outperforms the \textit{Random Baseline} across all metrics, showing the benefits of training even with noisy token labels derived from rationales. \textit{Multi-task} training further improves over \textit{Token-only} predictions
for almost all different metrics. \textit{Divergence Ranking} of contrastive instances yields the best results across the board. Also, on the auxiliary sentence-level task, the \textit{Multi-task} model matches the F1 as the standalone \textit{Divergence Ranking} model.

\begin{figure}[H]

    \begin{subfigure}{0.75\columnwidth}
      \hspace{0.9cm}
      \includegraphics[scale=0.2, left]{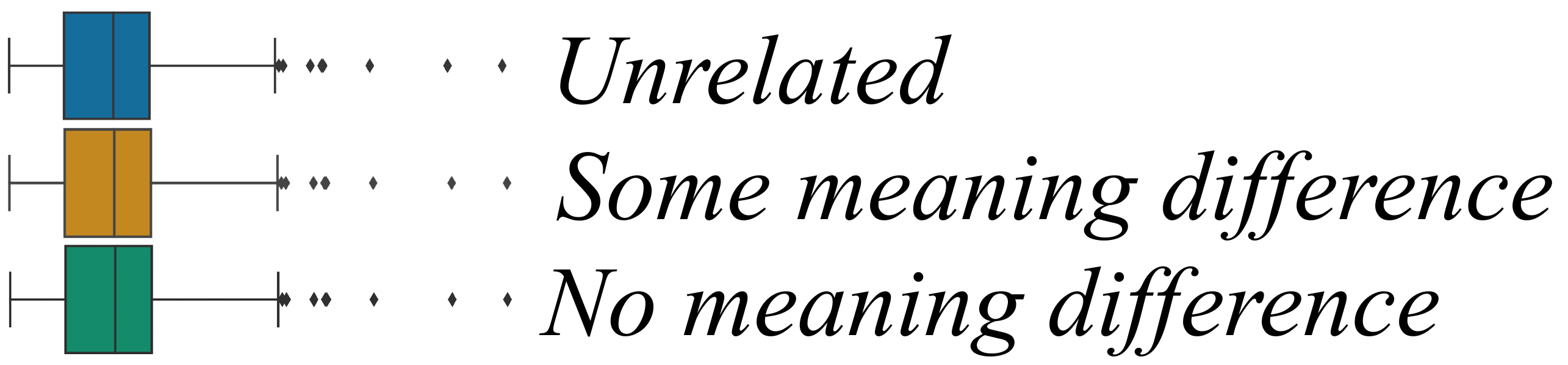}
    \end{subfigure}
    \vspace{0.2cm}

    \begin{subfigure}{0.72\columnwidth}
      \hspace{0.9cm}
      \includegraphics[width=6.5cm,height=3.8cm,center, left]{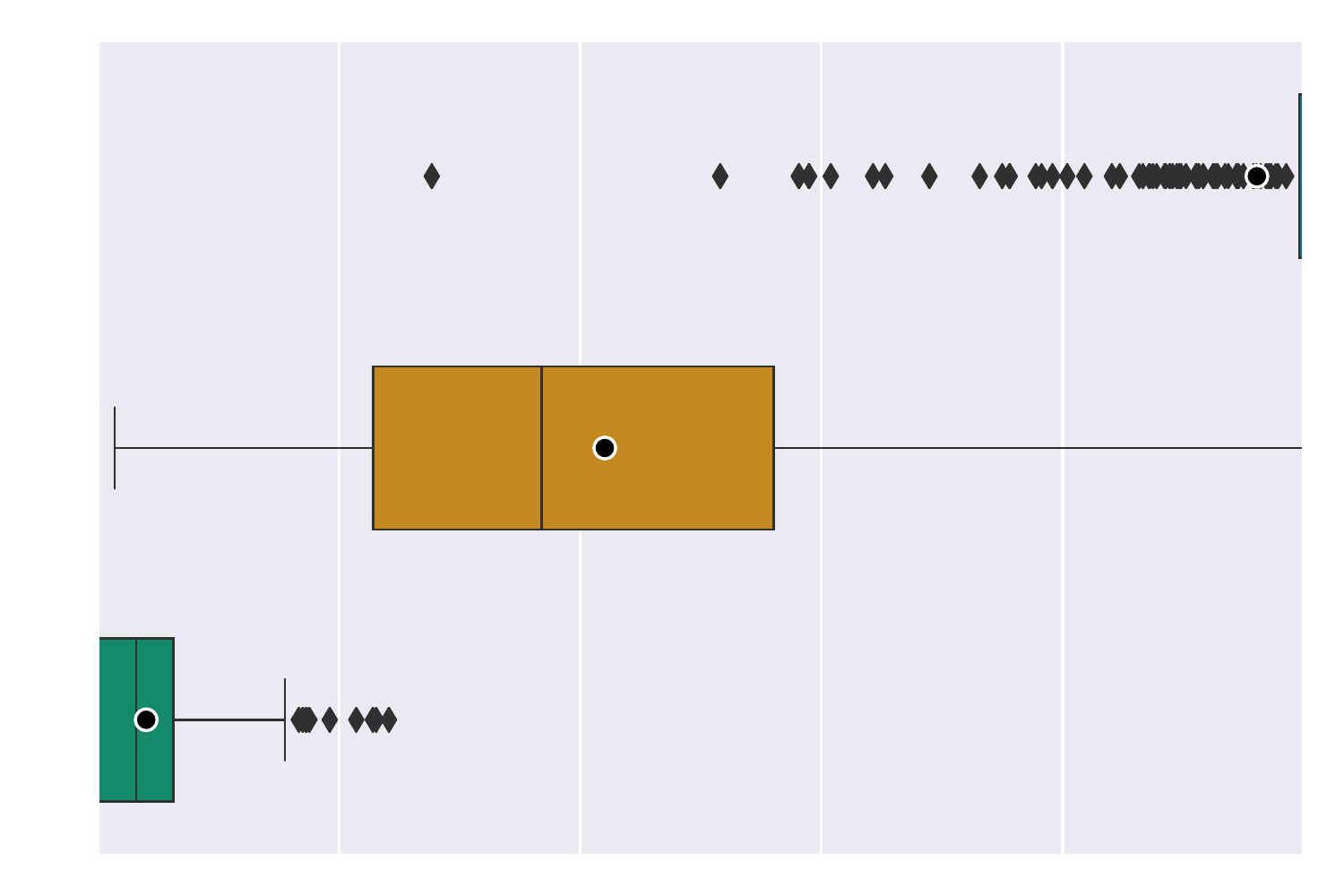}
    \end{subfigure}
    \vspace{-0.3cm}
    \caption*{ (a) \textit{Union}}

    \begin{subfigure}{0.77\columnwidth}
      \hspace{0.7cm}
      \includegraphics[width=6.8cm,height=3.8cm,center]{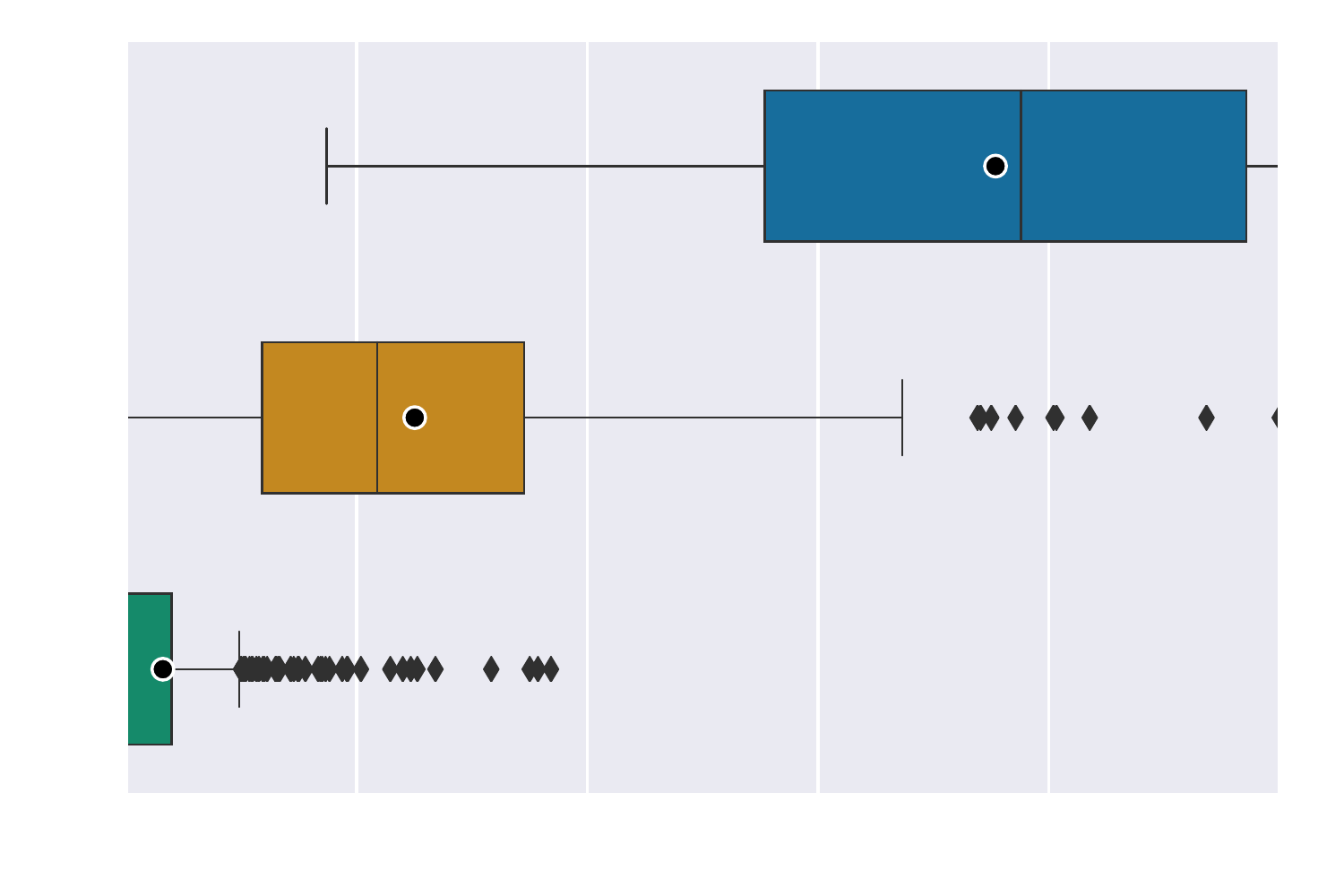}
    \end{subfigure}
    \vspace{-0.5cm}
    \caption*{(b) \textit{Pair-wise Union}}

    \begin{subfigure}{0.77\columnwidth}
      \hspace{0.7cm}
      \includegraphics[width=6.8cm,height=3.8cm,center, left]{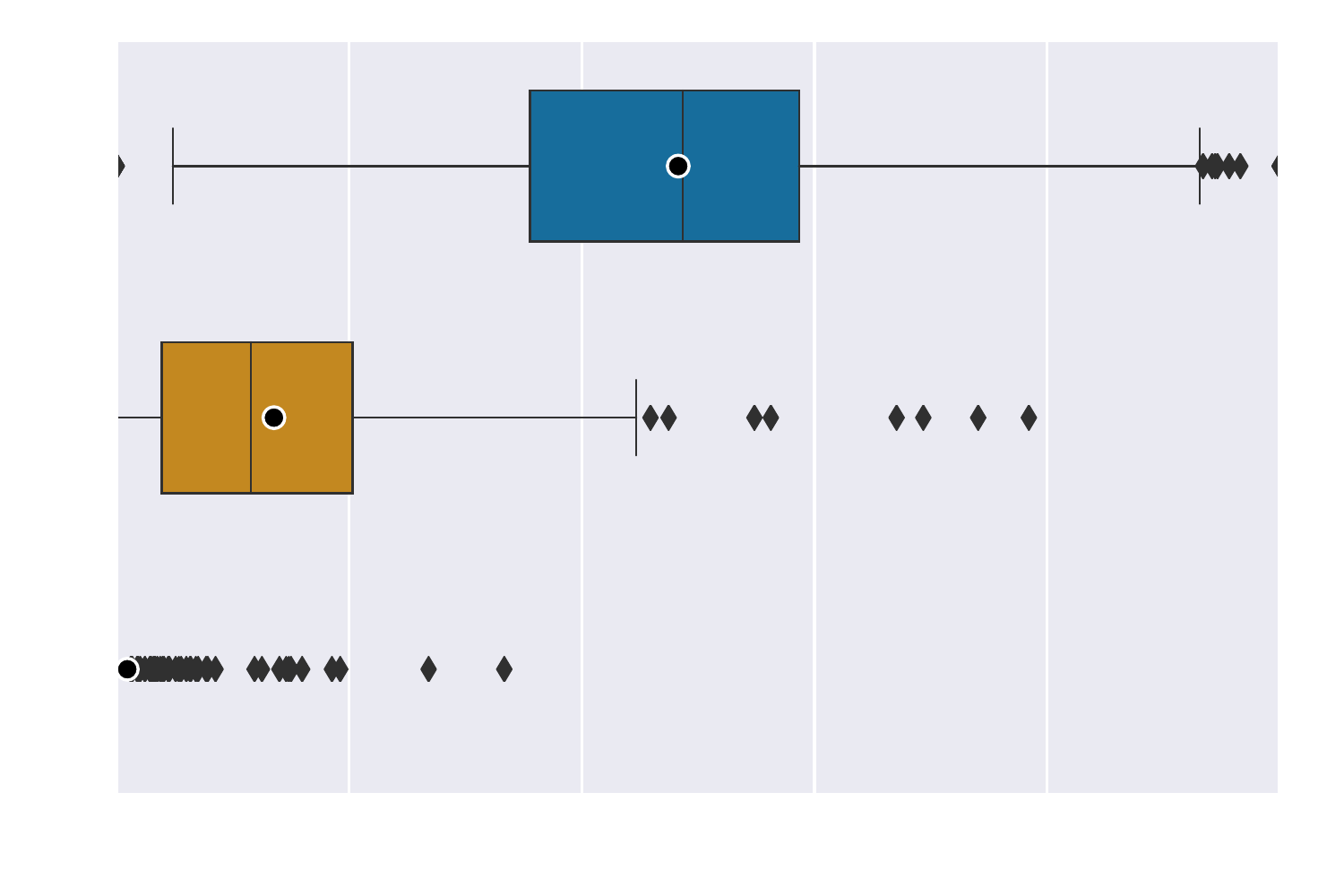}
    \end{subfigure}
    \vspace{-0.5cm}
     \caption*{(c) \textit{Intersection}}

    \begin{subfigure}{0.77\columnwidth}
      \hspace{0.7cm}
      \includegraphics[width=6.8cm,height=3.8cm,center, left]{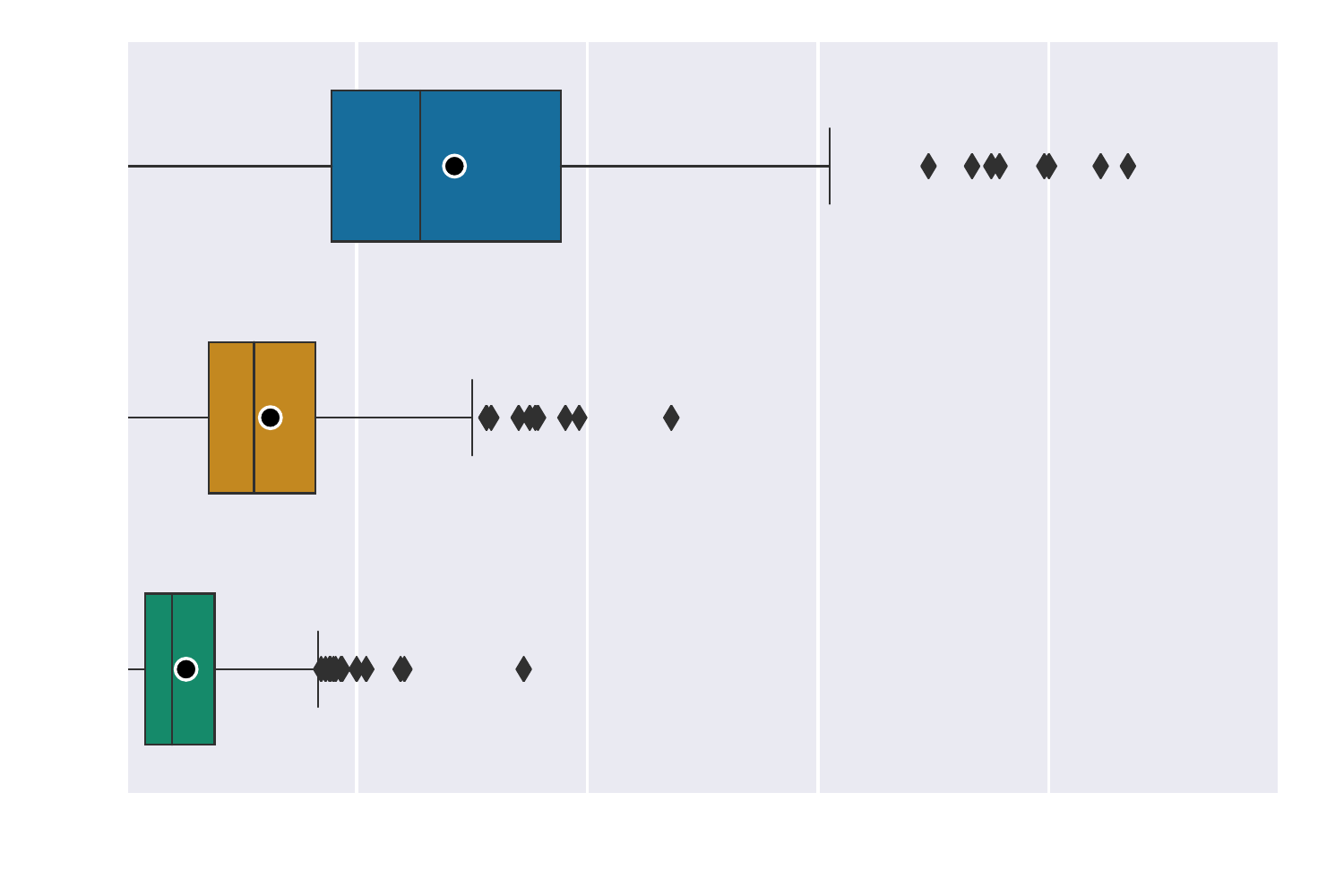}
    \end{subfigure}
    \vspace{-0.5cm}
    \caption*{(d) \textit{Token-only}}

    \begin{subfigure}{0.77\columnwidth}
      \hspace{0.7cm}
      \includegraphics[width=6.8cm,height=3.8cm,center]{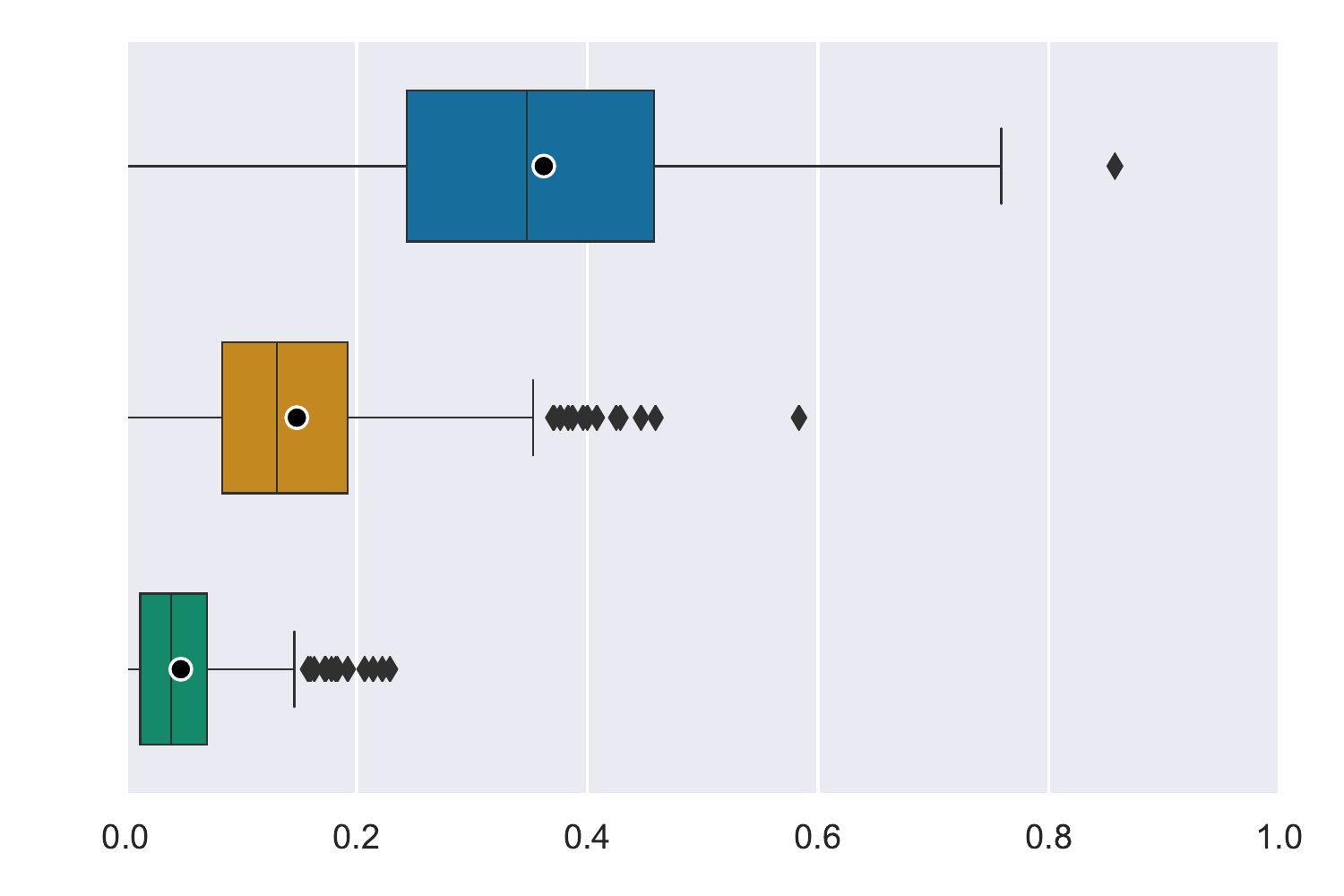}  
    \end{subfigure}
    \vspace{-0.3cm}
   \caption*{ (e) \textit{Divergence Ranking}}

\caption{Percentage distributions of \negat tokens in \dataset and \negat token predictions of two models: \textit{Divergence Ranking} makes more \negat predictions compared to the \textit{Token-only} model, enabling a better distinction between the divergent classes.}  \label{fig:precentage_token}

\end{figure}

\paragraph{From Token to Sentence Predictions} 
We compute the \% of \negat predictions within a sentence-pair. 
The multi-task model makes more \negat predictions for the divergent classes as its \% distribution 
is more skewed towards greater values~(Figure~\ref{fig:precentage_token} (d) vs.\ (e)). 
We then show that the \% of \negat predictions of 
the \textit{Divergence Ranking} model 
can be used as an indicator for distinguishing between divergences of different granularity: intuitively, a
sentence pair with more \negat tokens should map to a coarse-grained divergence at a sentence-level. Table \ref{tab:some_vs_uner} shows that thresholding the \% of \negat tokens could be an effective discrimination strategy, which we will explore further in future work. 

\begin{table}[H]
    \centering
    \scalebox{0.93}{
    \begin{tabular}{lccccccc}
    & \multicolumn{3}{c}{\textsc{un}} & \multicolumn{3}{c}{\textsc{sd}}\\
    \textbf{\negat\%} & \textbf{P} & \textbf{R} & \textbf{F1} & \textbf{P} & \textbf{R} & \textbf{F1} & \textbf{F1-all} \\
    $10$ & $48$ & $\mathbf{97}$ & $64$ & $66$ & $51$ & $57$ & $59$\\
    $20$ & $69$ & $84$ & $\mathbf{76}$ & $\mathbf{83}$ & $79$ & $81$ & $80$\\
    $30$ & $82$ & $63$ & $71$ & $81$ & $\mathbf{85}$ & $\mathbf{83}$ & $\mathbf{81}$ \\
    $40$ & $\mathbf{94}$ & $35$ & $51$ & $73$ & $84$ & $78$ & $75$ \\
    \end{tabular}}
    \caption{``Some meaning difference'' (\textsc{sd}) vs.\ ``Unrelated'' (\textsc{un}) classification based on $\%$ of \negat labels.\mc{the font looks really too small in this table.}}
    \label{tab:some_vs_uner}
\end{table}

\section{Related Work}

Our work is closely related to but distinct from the Semantic Textual Similarity (\textsc{sts}) task that measures the \textbf{degree of equivalence} in the underlying semantics of paired snippets of text~\cite{agirre-etal-2016-semeval-2016, cer-etal-2017-semeval}. Most commonly, state-of-the-art models address the \textsc{sts} task via interaction models that use alignment mechanisms to integrate word-level interactions in their final 
predictions~\cite{he-lin-2016-pairwise, parikh-etal-2016-decomposable} or via learning vector representations of sentences that are then compared using distance-based measures \cite{nie-bansal-2017-shortcut,conneau-etal-2017-supervised, cer-etal-2018-universal,reimers-gurevych-2019-sentence, yang-etal-2019-parameter}.

\section{Conclusion}

We show that explicitly considering diverse semantic divergence types benefits both the annotation and prediction of divergences between texts in different languages.

We contribute \dataset, a new dataset of WikiMatrix sentences-pairs in English and French, annotated with semantic divergence classes and token-level rationales that justify the sentence level annotation. $64\%$ of samples are annotated as divergent, and $40\%$ of samples contain fine-grained meaning divergences, confirming that divergences are too frequent to ignore even in parallel corpora. We show that these divergences can be detected by a m\textsc{bert} model fine-tuned without annotated samples, by learning to rank synthetic divergences of varying granularity. 

Inspired by the rationale-based annotation process, we show that predicting token-level and sentence-level divergences jointly is a promising direction for further distinguishing between coarser and finer-grained divergences.

\section{Acknowledgements}

We thank the anonymous reviewers and the \textsc{clip} lab at \textsc{umd} for helpful comments. This material is based upon work supported by the National Science Foundation under Award No.\ $1750695$. Any opinions, findings, and conclusions or recommendations expressed in this material are those of the authors and do not necessarily reflect the views of the National Science Foundation.

\bibliography{emnlp2020,refs}
\bibliographystyle{acl_natbib}

\appendix
\section{Appendices}\label{sec:appendix}

\subsection{Implementation Details}\label{app:implemantation_details}

\paragraph{Training setup} We employ the Adam optimizer with initial learning rate \(\eta= 2\mathrm{e}{-5}\), fine-tune for at most $5$ epochs, and use early-stopping to select the best model. We use a batch size of $32$ for experiments that do not use contrastive training and a batch size of $16$ for those using contrastive training to establish a fair comparison.

\paragraph{Model setup} All of our models are based on the ``Multilingual \textsc{bert}-base model'' consisting of: $12$-layers, $768$-hidden size, $12$-heads and $110$M parameters.    

\paragraph{Average Runtime \& Computing Infrastructure}
Each experiment is run on a single GeForce \textsc{gtx} $1080$ \textsc{gpu}. 
For experiments run on either a single type of divergence (e.g., \textit{Subtree Deletion}) or using \textit{Balanced} sampling, the average duration time is $\sim0.4$ hours. For \textit{Divergence Ranking} and \textit{Concatenation}, sampling methods, training takes $\sim2$ hours to complete.

\paragraph{Hyperparameter search on margin} We perform a grid search on the margin parameter for each experiment that employs contrastive training. We experiment with values $\{3,4,5,6,7,8\}$ and pick the one corresponding to the best Weighted-F1 score on a synthetic development set. Table~\ref{tab:validation-variance} shows mean and variance results on both the development and the \dataset dataset for different $\xi$ values. In general, we observe that our model's performance on \dataset is not sensitive to the margin's choice, as reflected by the small variances on the \dataset Weighted-F1.

\begin{table}[!ht]
    \centering
    \scalebox{0.79}{
    \begin{tabular}{lcccc}
    \textbf{Synthetic} & \textbf{Dev} & \textbf{\dataset} & \textbf{$\xi$*} & \textbf{Dev*}\\
    Phrase replacement    & $91.83_{\pm 1.14}$ & $78.60_{\pm 1.84}$ & $7$ & $93$ \\
    Subtree Deletion      & $93.67_{\pm 2.22}$ & $82.67_{\pm 0.22}$ & $8$ & $95$\\
    Lexical Substitution  & $91.50_{\pm 0.25}$ & $70.50_{\pm 1.58}$ & $5$ & $92$\\
    Balanced              & $87.67_{\pm 0.56}$ & $80.33_{\pm 0.56}$ & $5$ & $88$\\
    Concatenation         & $89.03_{\pm 0.50}$ & $79.51_{\pm 0.67}$ & $5$ & $90$ \\
    Divergence Ranking    & $77.80_{\pm 1.36}$ & $83.67_{\pm 0.22}$ & $5$ & $79$\\
    \end{tabular}}
    \caption{Average results of \modelname as a function of the number of hyperparameter trials for the margin value ($\xi$). The first row corresponds to the sampling method used for creating synthetic contrastive training examples. The second and third rows correspond to the mean/variance of Weighted-F1 results, measured on the development and the \dataset dataset, respectively. The fourth row describes the best value of the margin hyperparameter (\textbf{$\xi$*}) for each experiment, while the last row denotes the corresponding Weighted-F1 score on the development set.}
    \label{tab:validation-variance}
\end{table}

\subsection{Very Deep Pair-Wise Interaction baseline}\label{app:VDPWI}

We compare against the Very Deep Pair-Wise Interaction (\textsc{vdpwi}) model repurposed by \citet{Vyas2018IdentifyingSD} to identify cross-lingual semantic divergence vs.\ equivalence. We fine-tune m\textsc{bert} models on coarsely-defined semantic synthetic divergent pairs, similarly to the authors. We report results on two crowdsourced datasets, consisting of equivalence vs.\ divergence labels for $300$ sentence-pairs, drawn from the noisy OpenSubtitles and CommonCrawl corpora. The two evaluation datasets are available at: \url{https://github.com/yogarshi/SemDiverge/tree/master/dataset}.
\begin{table}[!ht]
    \scalebox{0.82}{
    \begin{tabular}{l@{\hskip 0.2in}ccc@{\hskip 0.2in}ccc}
     & \multicolumn{3}{c}{OpenSubtitles} &  \multicolumn{3}{c}{CommonCrawl}\\
    \textbf{Method} & \textbf{F1-} & \textbf{F1+} & \textbf{F1} & \textbf{F1-} & \textbf{F1+} & \textbf{F1} \\
    \citet{Vyas2018IdentifyingSD} & $78$ & $72$ & $77$ & $85$ & $73$ & $80$\\
    m\textsc{bert} & $\mathbf{81}$ & $\mathbf{76}$ & $\mathbf{79}$ & $\mathbf{87}$ & $\mathbf{76}$ & $\mathbf{83}$\\
    \end{tabular}}
    \caption{Performance comparison between m\textsc{bert} and \textsc{vdpwi} trained on coarsely-generated semantic divergences. We report F1 overall results (F1) and F1+/F1- scores for the two classes, on the crowdsourced OpenSubtitles and CommonCrawl datasets.}
    \label{tab:vdpwi}
\end{table}

Table~\ref{tab:vdpwi} presents results on the OpenSubtitles and CommonCrawl testbeds. We observe that m\textsc{bert} trained on similarly defined coarse divergences performs better than cross-lingual \textsc{vdpwi}.

\subsection{\dataset: Annotation Guidelines}\label{sec:annotation_guidelines}

Below we include the annotation guidelines given to participants:\\

``You are asked to compare the meaning of English and French text excerpts. You will be presented with one pair of texts at a time (about a sentence in English and a sentence in French). For each pair, you are asked to do the following:

\paragraph{1} Read the two sentences carefully.  Since the sentences are provided out of context,  your understanding of content should only rely on the information available in the sentences. There is no need to guess what additional information might be available in the documents the excerpts come from.

\paragraph{2} Highlight the text spans that convey different meaning in the two sentences. After highlighting a span of text, you will be asked to further characterize it as:

\begin{tabular}{p{0.82\columnwidth}}
\\
\textcolor{blue(pigment)}{\textbf{\textsc{added}}} \it the highlighted span corresponds to a piece of information that \textbf{does not exist} in the other sentence\\
\textcolor{blue(pigment)}{\textbf{\textsc{changed}}} \it the highlighted span corresponds to a piece of information that exists in the other sentence, but \textbf{their meaning is not the exact same} \\
\textcolor{blue(pigment)}{\textbf{\textsc{other}}} \it none of the above holds\\
\\
\end{tabular}

\noindent
You can highlight as many spans as needed. 
You can \textbf{optionally} provide an explanation for your assessment in the text form under the Notes section (e.g., literal translation of idiom)

\paragraph{3} Compare the meaning of the two sentences by picking one of the three classes:

\begin{tabular}{p{0.82\columnwidth}}
\\
\textcolor{bostonuniversityred}{\textbf{\textsc{unrelated}}} \it  The two sentences are completely unrelated or have a few words in common but convey unrelated information about them\\
\textcolor{bostonuniversityred}{\textbf{\textsc{some meaning difference}}} \it The two sentences convey \textbf{mostly the same information, except 
    differences for some details or nuances} (e.g., some information is added
    and/or missing on either or both sides; some English words have
    a more general or specific translation in French)   \\
\textcolor{bostonuniversityred}{\textbf{\textsc{no meaning difference}}} \it The two sentences have the \textbf{exact same meaning}''\\
\\
\end{tabular}

\subsection{Annotation Procedures}\label{sec:annotation_procedures}
We run $8$ online annotation sessions. Each session consists of $120$ instances, annotated by $3$ participants, and lasts about $2$ hours. Participants are allowed to take breaks during the process. Participants are rewarded with Amazon gift cards 
at a rate of \$$2$ per 10 examples, with bonuses of \$$5$ and \$$10$ for completing the first and additional sessions, respectively. 

\subsection{Annotated examples in \dataset}\label{sec:exampled_of_refresd}

Table~\ref{tab:refresd_examples} includes examples of annotated instances drawn from \dataset, corresponding to different levels of inter-annotator agreement.

{
\renewcommand{\arraystretch}{1.3}
\begin{table*}[!ht]
    \centering
    \scalebox{0.54}{
    \begin{tabular}{ll}

    \toprule[3pt]
    & \textbf{No meaning difference with \textit{high} sentence-level agreement and \textit{high} span overlap (n=3)}\\
    \addlinespace[0.2cm]
    \toprule[2pt]  

    \addlinespace[0.3cm]
    \\\textbf{EN} & The plan was revised in 1916 to concentrate the main US naval fleet in New England, and from there defend the US from the German navy.\\
    \addlinespace[0.2cm]
    \textbf{FR} &  Le plan fut révisé en 1916 pour concentrer le gros de la flotte navale américaine en Nouvelle-Angleterre, et à partir de là, défendre les États-Unis contre la marine allemande.\\
    
    \addlinespace[0.3cm]
    \\\toprule[3pt]
    & \textbf{Some meaning difference with \textit{high} sentence-level agreement and \textit{high} span overlap (n=3)}\\
     \addlinespace[0.2cm]
    \toprule[2pt]  
    
    \addlinespace[0.3cm]
    \\\textbf{EN} & After an intermediate period during which Stefano Piani edited the stories, in 2004 a major rework of the series went through.\\
    \addlinespace[0.2cm]
    \textbf{FR} & Après une période intermédiaire pendant laquelle Stefano Piani édita les histoires, une refonte majeure de la série fut faite en 2004 \colorbox{blue3}{\hz en réponse à une baisse notable des ventes.}\\
    
     \addlinespace[0.3cm]
    \\\toprule[3pt]
    & \textbf{Unrelated with \textit{high} sentence-level agreement and \textit{high} span overlap (n=3)}\\
     \addlinespace[0.2cm]
    \toprule[2pt]  
    
    \addlinespace[0.3cm]
    \\ \textbf{EN} &  \colorbox{blue1}{\hz To reduce vibration, all helicopters have rotor adjustments for height and weight.}\\
    \addlinespace[0.2cm]
    \textbf{FR} & \colorbox{blue2}{\hz En vol, le régime du compresseur Tous les compresseurs ont un taux de compression lié à la vitesse de rotation et au nombre d'étages.}\\
 
     \addlinespace[0.3cm]
     \\\toprule[3pt]
     & \textbf{No meaning difference with \textit{high} sentence-level agreement and \textit{high} span overlap (n=3)}\\
     \addlinespace[0.2cm]
    \toprule[2pt]  
 
     \addlinespace[0.3cm]
    \\\textbf{EN} & One \colorbox{blue1}{\hz can see} two sunflowers on the main façade and three smaller ones on the first floor above ground \colorbox{blue2}{\hz just} above the entrance arcade.\\
    \addlinespace[0.2cm]
    \textbf{FR} & On \colorbox{blue2}{\hz remarquera} deux tournesols sur la façade principale et trois plus petits au premier étage au-dessus des arcades d'entrée.\\
     
    \addlinespace[0.3cm]
    \\\toprule[3pt]
     & \textbf{Some meaning difference with \textit{high} sentence-level agreement and \textit{low} span overlap (n=3)}\\
    \addlinespace[0.2cm]
    \toprule[2pt]  
     
    \addlinespace[0.3cm]
    \\\textbf{EN} & On November 10, 2014, CTV ordered a fourth season \colorbox{blue1}{\hz of Saving Hope that consisted of} eighteen episodes, and premiered on September 24.\\
    \addlinespace[0.2cm]
    \textbf{FR} & Le 10 novembre 2014, CTV a \colorbox{blue1}{\hz renouvelé la série} pour une quatrième saison de 18 épisodes diffusée depuis le 24 \colorbox{blue1}{\hz septembre}\colorbox{blue2}{\hz 2015.}\\

    \addlinespace[0.3cm]
    \\\toprule[3pt]
     & \textbf{Unrelated with \textit{high} sentence-level agreement and \textit{low} span overlap (n=3)}\\
    \addlinespace[0.2cm]
    \toprule[2pt]  
    
    \addlinespace[0.3cm]
    \\\textbf{EN} & \colorbox{blue1}{\hz He} talks about \colorbox{blue2}{\hz Jay Gatsby, the most hopeful man}\colorbox{blue1}{ \hz he had ever met}.\\
    \addlinespace[0.2cm]
    \textbf{FR} & \colorbox{blue1}{\hz Il côtoie notamment}\colorbox{blue3}{\hz Giuseppe Meazza}\colorbox{blue2}{\hz qui}\colorbox{blue1}{\hz dira de lui}\colorbox{blue3}{\hz Il fut le joueur le plus fantastique que}\colorbox{blue2}{\hz j'aie eu l'occasion de voir.}\\
    
        \addlinespace[0.3cm]
    \\\toprule[3pt]
    & \textbf{No meaning difference with \textit{moderate} sentence-level agreement (n=2)}\\
    \addlinespace[0.2cm]
    \toprule[2pt]  
    
    \addlinespace[0.3cm]
    \\\textbf{EN} & Nine of these revised BB LMs were built by Ferrari in 1979, while a \colorbox{blue1}{\hz further refined} series of sixteen were built from 1980 to 1982.\\
    \addlinespace[0.2cm]
    \textbf{FR} & Neuf de ces BB LM révisées furent construites par Ferrari en 1979, tandis qu'une série de seize autres furent construite entre 1980 et 1982.\\
    
        \addlinespace[0.3cm]
    \\\toprule[3pt]
    & \textbf{Some meaning difference with \textit{moderate} sentence-level agreement (n=2)}\\
    \addlinespace[0.2cm]
    \toprule[2pt]     
    
    \addlinespace[0.3cm]
    \\\textbf{EN} & From 1479, the Counts of Foix became \colorbox{blue1}{\hz Kings} of Navarre and \colorbox{blue1}{\hz the last of them}, \colorbox{blue1}{\hz made} Henri IV of France, annexed his Pyrrenean lands to France.\\
    \addlinespace[0.2cm]
    \textbf{FR} & À partir de 1479, le comte de Foix \colorbox{blue2}{\hz devient} roi de Navarre et le dernier d’entre eux, \colorbox{blue1}{\hz devenu} Henri IV, \colorbox{blue2}{\hz roi de}\colorbox{blue1}{\hz France }\colorbox{blue3}{\hz en 1607,}\colorbox{blue1}{\hz annexe} ses terres pyrénéennes à la France.\\
    
        \addlinespace[0.3cm]
    \\\toprule[3pt]
    & \textbf{Unrelated difference with \textit{moderate} sentence-level agreement (n=2)}\\
    \addlinespace[0.2cm]
    \toprule[2pt]    
    
    \addlinespace[0.3cm]
    \\\textbf{EN} & \colorbox{blue3}{\hz The operating principle was the same as that used in the Model 07/12 Schwarzlose}\colorbox{blue2}{\hz machine gun}\colorbox{blue1}{\hz used} by Austria-Hungary during the First World War.\\
    \addlinespace[0.2cm]
    \textbf{FR} & \colorbox{blue3}{\hz Le Skoda 100 mm modèle 1916}\colorbox{blue2}{\hz était un obusier de montagne}\colorbox{blue1}{\hz utilisé} par l'Autriche-Hongrie pendant la Première Guerre mondiale.\\
        \addlinespace[0.3cm]

    \\\toprule[3pt]

    \end{tabular}} 
    \caption{\dataset examples, corresponding to different levels of agreement between annotators. $n$ denotes the number of annotators who voted for the sentence-level majority class; disagreements span closely related classes.} 
  \label{tab:refresd_examples}
\end{table*}
}

\subsection{Token predictions of \modelname}\label{app:token_predictions_example}

Table~\ref{tab:token_output_examples} shows randomly selected instances from \dataset along with token tags predicted by our best performing system
(\textit{Divergence Ranking}).

{
\renewcommand{\arraystretch}{1}
\begin{table*}[!ht]
    \centering
    \scalebox{0.54}{
    \begin{tabular}{ll}

    \toprule[3pt]
  \addlinespace[1cm]

\multirow{2}{*}{\textbf{EN}} & He \colorbox{blue1}{\hz joined the}\colorbox{blue2}{\hz Munich State Opera}\colorbox{blue3}{\hz in 1954,} where he created the role of Johannes Kepler in Hindemith's Die Harmonie der Welt (1957).\\

& He \colorbox{green}{\hz joined the} Munich \colorbox{green}{\hz State Opera in 1954, where} he created the role of Johannes Kepler in Hindemith's Die Harmonie der Welt (1957).\\

 \addlinespace[0.2cm]

\multirow{2}{*}{\textbf{FR}} & \colorbox{blue1}{\hz Il crée à Munich,} le rôle de Johannes Kepler dans Die Harmonie der Welt de Paul Hindemith en 1957.\\

& Il crée à Munich, le rôle de Johannes Kepler dans Die Harmonie der Welt de Paul Hindemith en 1957.\\

 \addlinespace[1cm]
 \hline
 \addlinespace[1cm]

\multirow{2}{*}{\textbf{EN}} & \colorbox{blue2}{\hz He experimented with silk vests}\colorbox{blue1}{\hz resembling medieval gambesons,} which used 18 to 30 layers\colorbox{blue1}{\hz of}\colorbox{blue2}{\hz silk fabric}\colorbox{blue1}{\hz to protect}\colorbox{blue2}{\hz the wearers from penetration.}\\

& He \colorbox{green}{\hz experimented} with silk vests resembling medieval gambesons, which used 18 to 30 layers of \colorbox{green}{\hz silk fabric} to protect the \colorbox{green}{\hz wearers} from \colorbox{green}{\hz penetration.}\\

\addlinespace[0.2cm]

\multirow{2}{*}{\textbf{FR}} & \colorbox{blue1}{\hz Ils ressemblaient aux jaques, vêtements matelassés médiévaux} constitués de 18 à 30 couches de \colorbox{blue1}{\hz vêtements} afin \colorbox{blue1}{\hz d'offrir une protection}\colorbox{blue2}{\hz maximale}\colorbox{blue3}{\hz contre les flèches.}\\

& Ils ressemblaient aux jaques, vêtements matelassés médiévaux constitués de 18 à 30 couches de \colorbox{green}{\hz vêtements} afin d'offrir une \colorbox{green}{\hz protection maximale} contre les \colorbox{green}{\hz flèches.} \\

 \addlinespace[1cm]
 \hline
 \addlinespace[1cm]

\multirow{2}{*}{\textbf{EN}} & \colorbox{blue3}{\hz Even though this made Armenia a client kingdom,} various contemporary Roman sources thought that Nero had de facto ceded Armenia to the Parthian Empire.\\

& Even though this made Armenia \colorbox{green}{\hz a client kingdom}, various contemporary Roman sources thought that Nero had de facto ceded Armenia to the Parthian \colorbox{green}{\hz Empire}.\\

 \addlinespace[0.2cm]

\multirow{2}{*}{\textbf{FR}} & Plusieurs sources romaines contemporaines n'en ont pas moins considéré que Néron a ainsi de facto cédé l'Arménie aux Parthes.\\

& Plusieurs sources romaines contemporaines n'en ont pas moins considéré que Néron a ainsi de facto cédé l'Arménie aux Parthes.\\

 \addlinespace[1cm]
 \hline
 \addlinespace[1cm]
 
\multirow{2}{*}{\textbf{EN}} &The Photo League was a cooperative of photographers in New York who banded together around a range of common social and creative causes.\\

& The Photo League was a cooperative of photographers in New York who banded together around a \colorbox{green}{\hz range} of common social and creative \colorbox{green}{\hz causes}.\\

 \addlinespace[0.2cm]

\multirow{2}{*}{\textbf{FR}} & La Photo League était un groupement de photographes \colorbox{blue3}{\hz amateurs et professionnels}\colorbox{blue1}{\hz réuni} à New York autour d'objectifs communs de nature sociale et créative.\\

& La Photo League était un groupement de photographes \colorbox{green}{\hz amateurs et professionnels} réuni à New York autour \colorbox{green}{\hz d'objectifs} communs de \colorbox{green}{\hz nature} sociale et créative.\\

 \addlinespace[1cm]
 \hline
 \addlinespace[1cm]
 
\multirow{2}{*}{\textbf{EN}} & \colorbox{blue1}{\hz She made}\colorbox{blue2}{\hz a}\colorbox{blue3}{\hz courtesy call}\colorbox{blue1}{\hz to} the Hawaiian Islands at the end of the year \colorbox{blue1}{\hz and}\colorbox{blue3}{\hz proceeded thence}\colorbox{blue1}{\hz to} Puget Sound \colorbox{blue3}{\hz where she arrived} on 2 February 1852.\\

& \colorbox{green}{\hz She} made a \colorbox{green}{\hz courtesy call} to the Hawaiian Islands at the end of the year and proceeded thence to Puget Sound where \colorbox{green}{\hz she} arrived on 2 February 1852.\\

 \addlinespace[0.2cm]

\multirow{2}{*}{\textbf{FR}} & \colorbox{blue1}{\hz Il fait}\colorbox{blue2}{\hz une}\colorbox{blue3}{\hz escale} aux îles Hawaï à la fin de l'année, \colorbox{blue1}{\hz au} Puget Sound, le 2 février 1852.\\

& \colorbox{green}{\hz Il} fait une escale aux îles Hawaï à la fin de l'année, au Puget Sound, le 2 février 1852.\\

 \addlinespace[1cm]
 \hline
 \addlinespace[1cm]

\multirow{2}{*}{\textbf{EN}} & Recognizing Nishikaichi and his plane as Japanese, Kaleohano thought it prudent to \colorbox{blue1}{\hz relieve} the pilot of his pistol and papers \colorbox{blue3}{\hz before the dazed airman could react.}\\

& Recognizing Nishikaichi and his plane as Japanese, Kaleohano thought it prudent to \colorbox{green}{\hz relieve} the pilot of his pistol and papers \colorbox{green}{\hz before the dazed airman could react}.\\

 \addlinespace[0.2cm]

\multirow{2}{*}{\textbf{FR}} & Reconnaissant Nishikaichi et son avion comme étant japonais, Kaleohano pensa qu'il serait prudent de \colorbox{blue1}{confisquer} au pilote son pistolet et ses documents.\\

& Reconnaissant Nishikaichi et son avion comme étant japonais, Kaleohano pensa qu 'il serait prudent de \colorbox{green}{confisquer} au pilote son pistolet et ses documents.\\

 \addlinespace[1cm]
 \hline
 \addlinespace[1cm]

\multirow{2}{*}{\textbf{EN}} & \colorbox{blue3}{\hz At the same time ,}\colorbox{blue1}{\hz the mortality rate increased slightly from 8.9 per 1,000 inhabitants in 1981 to 9.6 per 1,000 inhabitants in 2003.}\\

& At the same time, the mortality rate increased slightly from 8.9 per \colorbox{green}{\hz 1,000 inhabitants in 1981} to 9.6 per \colorbox{green}{\hz 1,000 inhabitants in 2003}.\\

 \addlinespace[0.2cm]

\multirow{2}{*}{\textbf{FR}} & \colorbox{blue1}{\hz Le taux de mortalité}\colorbox{blue2}{\hz est}\colorbox{blue3}{\hz quant à lui passé de 11,8 \% sur la période 1968-1975 à 9,1 \% sur la période 1999-2009 .}\\

& Le taux de mortalité est quant à lui passé de 11,8 \% sur la \colorbox{green}{\hz période 1968-1975} à 9,1 \% sur la \colorbox{green}{\hz période 1999-2009.}\\

 \addlinespace[1cm]
 \hline
 \addlinespace[1cm]
 
\multirow{2}{*}{\textbf{EN}} & \colorbox{blue1}{\hz They called for a state convention on September 17 in Columbia to nominate a statewide ticket.}\\
& They called for a \colorbox{green}{\hz state} convention \colorbox{green}{\hz on September 17 in Columbia to nominate a statewide ticket.}\\

 \addlinespace[0.2cm]

\multirow{2}{*}{\textbf{FR}} & \colorbox{blue3}{\hz Un décret de la Convention du 28 avril 1794 ordonna que son nom fût inscrit sur une colonne de marbre au Panthéon.}\\

& Un \colorbox{green}{\hz décret} de la Convention \colorbox{green}{\hz du 28 avril 1794 ordonna} que \colorbox{green}{\hz son nom} fût \colorbox{green}{\hz inscrit sur une colonne de marbre au Panthéon.}\\

 \addlinespace[1cm]
 \hline
 \addlinespace[1cm]

\multirow{2}{*}{\textbf{EN}} & \colorbox{blue1}{\hz His plants are still in the apartment and the two take all of the plants with them back to their place.}\\
& His \colorbox{green}{\hz plants} are \colorbox{green}{\hz still} in the \colorbox{green}{\hz apartment} and the two \colorbox{green}{\hz take} all of the \colorbox{green}{\hz plants} with \colorbox{green}{\hz them} back to \colorbox{green}{\hz their place.}\\

 \addlinespace[0.2cm]

\multirow{2}{*}{\textbf{FR}} & \colorbox{blue3}{\hz Il reste}\colorbox{blue2}{\hz donc chez lui et les deux sœurs}\colorbox{blue3}{\hz s'occupent du show toutes seules.}\\

& Il reste donc chez lui et les deux \colorbox{green}{\hz sœurs} s'occupent du \colorbox{green}{\hz show toutes seules.}\\

\addlinespace[1cm]

\toprule[3pt]

    \end{tabular}}
    \caption{\dataset examples, along with \modelname's predictions. Tokens highlighted with green color correspond to \negat predictions of \modelname (second sentence). Tokens highlighted with red colors correspond to gold-standard labels of divergence provided by annotators (first sentence). The red color intensity denotes the degree of agreement across three annotators (darker color denotes higher agreement).}
    \label{tab:token_output_examples}
\end{table*}}

\subsection{Results on synthetic development sets}\label{sec:dev_suppl}

Tables \ref{tab:sentence-dev} and \ref{tab:token-synthetic}  report results on development sets for each experiment included in Tables \ref{tab:sentence_divergences_single} and \ref{tab:token-level-some_meanining-only}, respectively. 

{
\setlength{\tabcolsep}{10pt}{
\renewcommand{\arraystretch}{1}
\begin{table*}[!ht]
    \centering
    \scalebox{0.84}{
    \begin{tabular}{llcrrrrrrrrr}
    \toprule[1.4pt]
     & & & \multicolumn{3}{c}{\textit{Equivalents}}& \multicolumn{3}{c}{\textit{Divergents}} & \multicolumn{3}{c}{\textit{All}} \\
    \cmidrule(lr){4-6}\cmidrule(lr){7-9}\cmidrule(lr){10-12}
    \textbf{Divergent} & \textbf{Loss} & \textbf{Contrastive} & \textbf{P+} & \textbf{R+} & \textbf{F1+} & \textbf{P-} & \textbf{R-} & \textbf{F1-} &  \textbf{P} &  \textbf{R} & \textbf{F1}\\
    \addlinespace[0.2cm]
    \toprule[1pt]
    
    \addlinespace[0.2cm]
    \multirow{3}{*}{\textit{Phrase Replacement}} & \multirow{2}{*}{CE} & \xmark & $92$ & $97$ & $94$ & $96$ & $92$ & $94$ & $94$ & $94$ & $94$  \\
     & & \cmark & $92$ & $97$ & $94$ & $97$ & $91$ & $94$ & $94$ & $94$ & $94$  \\
     & Margin & \cmark & $91$ & $95$ & $93$ & $95$ & $91$ & $93$ & $93$ & $93$ & $93$ \\ 

    \addlinespace[0.2cm]
    \multirow{3}{*}{\textit{Subtree Deletion}} & \multirow{2}{*}{CE} & \xmark & $93$ & $97$ & $95$ & $97$ & $93$ & $95$ & $95$ & $95$ & $95$      \\
    & & \cmark & $94$ & $97$ & $96$ & $97$ & $94$ & $96$ & $96$ & $96$ & $96$ \\
    & Margin  & \cmark & $93$ & $97$ & $95$ & $97$ & $93$ & $95$ & $95$ & $95$ & $95$ \\

    \addlinespace[0.2cm]
    \multirow{3}{*}{\textit{Lexical Substitution}} & \multirow{2}{*}{CE} & \xmark &  $93$ & $94$ & $94$ & $94$ & $93$ & $93$ & $94$ & $94$ & $93$   \\
    & & \cmark & $95$ & $93$ & $94$ & $94$ & $95$ & $94$ & $94$ & $94$ & $94$ \\
    & Margin & \cmark & $91$ & $94$ & $93$ & $94$ & $91$ & $92$ & $92$ & $92$ & $92$  \\
    
    \addlinespace[0.4cm]
    \hline
    
    \addlinespace[0.2cm]
    \multirow{3}{*}{\textit{Balanced}} & \multirow{2}{*}{CE} & \xmark & $90$ & $96$ & $92$ & $95$ & $89$ & $92$ & $92$ & $92$ & $92$  \\
    & & \cmark & $90$ & $94$ & $92$ & $94$ & $90$ & $92$ & $92$ & $92$ & $92$ \\
    & Margin  & \cmark & $85$ & $93$ & $89$ & $92$ & $84$ & $88$ & $89$ & $88$ & $88$  \\
    
    \addlinespace[0.2cm]
    \multirow{3}{*}{\textit{Concatenation}} & \multirow{2}{*}{CE} & \xmark & $92$ & $90$ & $91$ & $90$ & $92$ & $91$ & $91$ & $91$ & $91$   \\
    & & \cmark & $82$ & $89$ & $86$ & $97$ & $95$ & $96$ & $94$ & $94$ & $94$ \\
    & Margin & \cmark & $89$ & $92$ & $90$ & $91$ & $88$ & $90$ & $90$ & $90$ & $90$ \\
  
    \addlinespace[0.4cm]
    \textit{Divergence Ranking} & Margin & \cmark & $72$ & $96$ & $82$ & $94$ & $63$ & $75$ & $83$ & $79$ & $79$\\
    \addlinespace[0.2cm]
    \hline
    \toprule[1.4pt]
    \end{tabular}}
    \caption{Evaluation on \textbf{synthetic development sets}. 
     We report Precision (P), Recall (R), and F1 for the equivalent (+) and
     divergent (-) classes separately and both classes (All). 
     Each model uses a development set that includes divergent types used during training. \textit{Divergence Ranking} yields lower performance on the synthetic development set than \dataset, reflecting the mismatch between the nature of synthetics samples vs.\ divergences in \dataset.}
    \label{tab:sentence-dev}
\end{table*}
}} 

\newpage

{ 
\setlength{\tabcolsep}{10pt}{
\renewcommand{\arraystretch}{1}
\begin{table*}[!ht]
    \centering
    \scalebox{1}{
    \begin{tabular}{lcccc}
    \toprule[1.4pt]
    Model & Multi-task  & F1-\pos & F1-\negat & F1-Mul\\
    \toprule[0.5pt]
    \textit{Token-only} &  & $99$ & $88$ & $87$\\
    \textit{Balanced} & \cmark & $98$ & $71$ & $70$\\
    \textit{Concatenation} & \cmark& 	$98$ & $71$  & $70$\\
    \textit{Divergence Ranking} & \cmark &	$98$ & $75$ & $74$\\

    \addlinespace[0.2cm]
    \toprule[1.4pt]
    \end{tabular}}
    \caption{Evaluation of token tagging models on \textbf{synthetic test sets}. We report Precision (P), Recall (R), and F1 scores for each class. F1-Mul corresponds to the product of individual F1 scores. The model's performance on synthetic test sets is always better than the one reported on \dataset, reflecting the mismatch between the noisy training samples and the real divergences found in \dataset.}
    \label{tab:token-synthetic}
\end{table*}
}}

\end{document}